%% file: ms.tex
\documentclass[final]{acmsiggraph}
\usepackage[utf8x]{inputenc}
\usepackage{amsfonts}
\usepackage{xspace}
\usepackage{microtype}

\input{defs}

\title{StreetStyle: Exploring world-wide clothing styles from millions
  of photos}

\author{Kevin Matzen \thanks{e-mail:kmatzen@cs.cornell.edu} \and Kavita Bala \thanks{e-mail:kb@cs.cornell.edu} \and Noah Snavely \thanks{e-mail:snavely@cs.cornell.edu}}
\affiliation{Cornell University}

\pdfauthor{Kevin Matzen}


\begin{document}

 \teaser{
   \includegraphics[width=\textwidth]{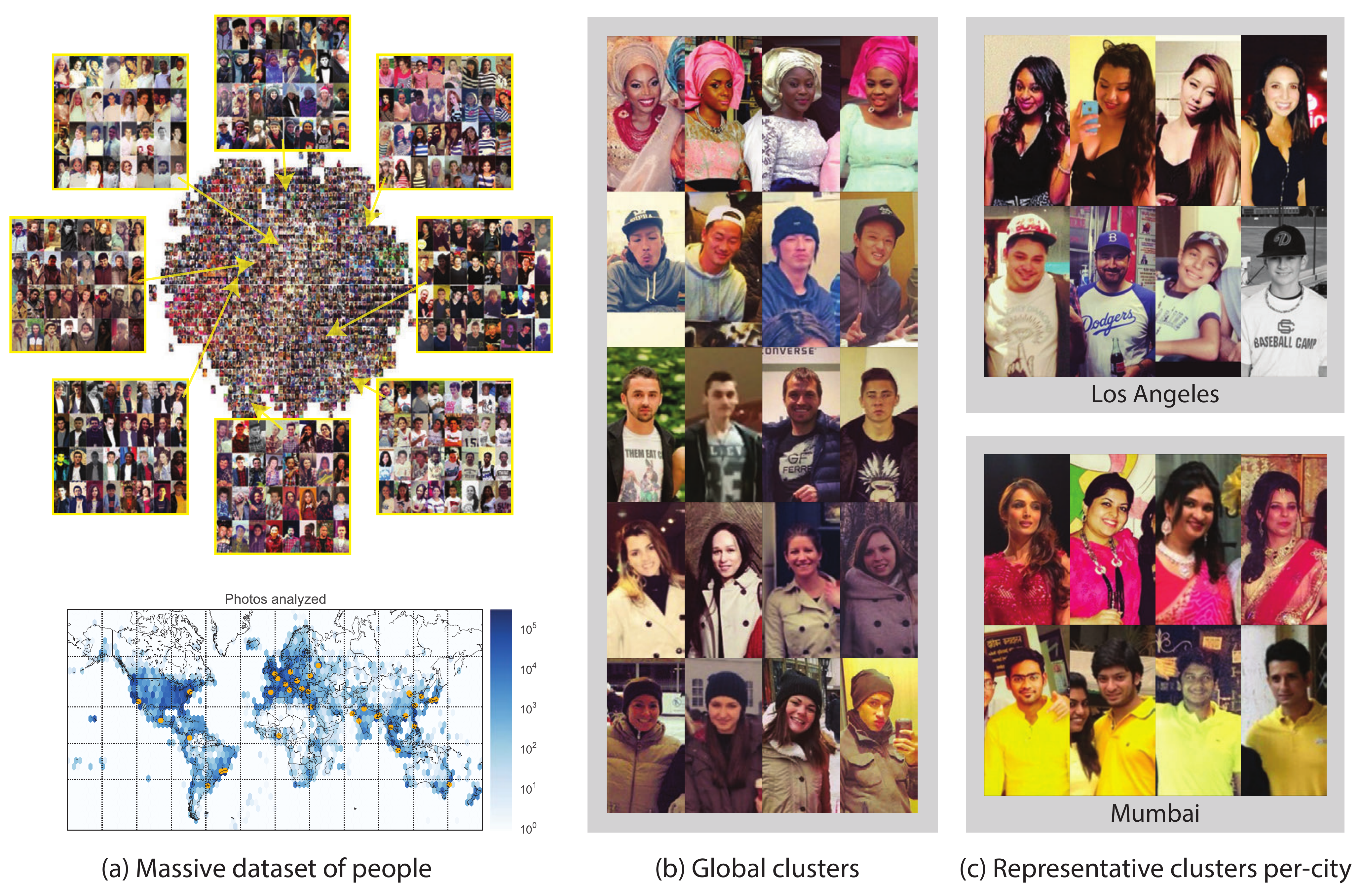}
   \caption{{\bf Extracting and measuring clothing style from Internet
       photos at scale.}  (a) We apply deep learning methods to learn
     to extract fashion attributes from images and create a visual
     embedding of clothing style. We use this embedding to analyze
     millions of Instagram photos of people sampled worldwide, in
     order to study spatio-temporal trends in clothing around the
     globe. (b) Further, using our embedding, we can cluster images to
     produce a global set of representative styles, from which we can
     (c) use temporal and geo-spatial statistics to generate concise
     visual depictions of what makes clothing unique in each city
     versus the rest.}
\label{fig:teaser}}

\maketitle

\input{streetstyle/abstract}

\input{streetstyle/intro}
\input{streetstyle/related}
\input{streetstyle/data}
\input{streetstyle/method}
\input{streetstyle/analysis}
\input{streetstyle/conclusion}

\bibliographystyle{acmsiggraph}
\bibliography{all}
\end{document}

%% file: defs.tex
\usepackage{color}

\definecolor{kevin_color}{rgb}{0,0.5,1}
\definecolor{kb_color}{rgb}{0.2,.64,0}
\definecolor{noah_color}{rgb}{1,0,1}
\definecolor{kb_color_mark}{rgb}{0.2,.9,0}
\iffalse
\newcommand{\kevin}[1]{\textsf{\textcolor{kevin_color}{[{\bf Kevin says}: #1]}}}
\newcommand{\kb}[1]{\textsf{\textcolor{kb_color}{[{\bf KB says}: #1]}}}
\newcommand{\noah}[1]{\textsf{\textcolor{noah_color}{[{\bf Noah says}: #1]}}}

\else
\newcommand{\kevin}[1]{}
\newcommand{\kb}[1]{}
\newcommand{\noah}[1]{}

\fi

\newcommand{\etal}{{\em et~al.}}

\newenvironment{packed_enum}{
\begin{enumerate}
  \setlength{\itemsep}{1pt}
  \setlength{\parskip}{2pt}
  \setlength{\parsep}{0pt}
}{\end{enumerate}}

\newenvironment{packed_item}{
\begin{itemize}
  \setlength{\itemsep}{1pt}
  \setlength{\parskip}{2pt}
  \setlength{\parsep}{0pt}
}{\end{itemize}}

\newcommand{\leaveout}[1]{}

\newcommand{\percent}{\%}
\newcommand{\StreetStyle}{{\sc{StreetStyle-27k}}\xspace}

%% file: streetstyle/abstract.tex
\begin{abstract}
Each day billions of photographs are uploaded to photo-sharing
services and social media platforms. These images are packed with
information about how people live around the world. In this paper we
exploit this rich trove of data to understand fashion and style trends
worldwide.  We present a framework for {\em visual discovery} at
scale, analyzing clothing and fashion across millions of images of
people around the world and spanning several years.  We introduce a
large-scale dataset of photos of people annotated with
clothing attributes, and use this dataset to train attribute
classifiers via deep learning. We also present a method for
discovering visually consistent {\em style clusters} that capture
useful visual correlations in this massive dataset. Using these tools,
we analyze millions of photos to derive visual insight, producing a
first-of-its-kind analysis of global and per-city fashion choices and
spatio-temporal trends.
\end{abstract}

%% file: streetstyle/intro.tex
\section{Introduction} \label{sec:streetstyle_intro}

Our age of big data presents us with the compelling opportunity to use
all available information to measure the world in ways that were never
before possible. Large amounts of data---for instance, OCRed scans of
centuries worth of books---coupled with tools for exploring this data
have yielded powerful new mechanisms for scientists to study our
history, culture, and behavior~\cite{michel:science:2010}. This
opportunity is magnified by the massive scale at which humans are
generating cultural artifacts on social media, and by the increasing
power of machine learning techniques.  For instance, by applying
natural language processing to millions of Twitter messages, we can
discover relationships between time of day and mood that leverage
sample sizes much larger than those of any traditional
study~\cite{golder:science:2011}.

To date, most of these new kinds of big data analyses have been
limited to structured data, such as user interactions on social
networks, or to textual data, such as books and tweets. However, a
tremendous cache of unstructured visual information about our world is
locked in images, particularly in images of people, including the
billions of photos uploaded to photo-sharing services each
day. Imagine a future anthropologist with access to trillions of
photos of people---taken over centuries and across the world---and
equipped with effective tools for analyzing these photos to derive
insights. What kinds of new questions can be answered?  This problem
area of {\em data-driven visual discovery} is still new, but is
beginning to gain attention in computer vision and
graphics~\cite{doersch:siggraph:2012,wang:iccvw:2013,zhu:siggraph:2014,ginosar:iccvw:2015,gebru:arxiv:2017}.
Our work takes a step towards this vision by analyzing geo-spatial
trends in fashion style across tens of millions of images from social
media.

In this paper, we focus on clothing, as it is a critical aspect of the
visual world and of our daily lives. Individuals make fashion choices
based on many factors, including geography, weather, culture, and
personal preference.  The ability to analyze and predict trends in
fashion is valuable for many applications, including analytics for
fashion designers, retailers, advertisers, and manufacturers. This
kind of analysis is currently done manually by analysts by, for
instance, collecting and inspecting photographs from relevant
locations and times.

We aim to extract meaningful insights about the geo-spatial and
temporal distributions of clothing, fashion, and style around the
world through social media analysis at scale.  We do so through the
combination of (1) millions of photos spanning the world---photos
uploaded to social media services by everyday users, (2) a new
dataset, \StreetStyle, consisting of a subset of these images
annotated with fashion attributes, and (3) powerful machine learning
methods based on deep learning that leverage our dataset.  We explore
two kinds of machine learning methods: supervised learning methods
that are trained on our annotated dataset to predict clothing
attributes in new images, followed by unsupervised clustering methods
that can automatically detect visual correlations in our data (such as
particular types of headwear, as in Figure~\ref{fig:teaser}(b), top
row).
These machine learning methods allow us to measure clothing features
across millions of photos, and then to use these measurements to
produce analyses in the form of trend reports and map-based
visualizations, enabling new types of visual insight. For instance,
our framework can help answer questions such as:
\begin{packed_item}
\item How is the frequency of scarf use in the US changing over time?
  (Figure~\ref{fig:streetstyle_scarf})
\item What styles are most specific to particular regions of the world
  or time of the year? Conversely, which styles are popular across the
  world? (Figures~\ref{fig:streetstyle_rank_cities} and
  \ref{fig:sorted_clusters})
\item For a given city, such as Los Angeles, what styles are most
  characteristic of that city (popular in LA, but rare elsewhere)?
  (Figures~\ref{fig:teaser}(c) and \ref{fig:streetstyle_rank_cities}.
\end{packed_item}
Our approach also demonstrates the utility of machine learning methods
for vastly simplifying the process of making real-world measurements
from large-scale visual data.

In summary, our work makes the following contributions:
\begin{packed_enum}
\item \StreetStyle, an annotated dataset of people containing 27K
  images, each with 12 clothing attributes, to be made publicly
  available,
\item a methodology for analyzing millions of photos to produce a
  visual clothing embedding (shown in Figure~\ref{fig:teaser}(a)),
  then using this embedding to predict clothing attributes in new
  photos,
\item the use of unsupervised clustering methods to automatically
  predict visual correlations between clothing attributes (in the form
  of {\em style clusters}, Figure~\ref{fig:teaser}(b)), and
\item a first-of-its-kind analysis of global and per-city fashion
  choices and trends using our machine learning methods.
\end{packed_enum}
We provide additional visualizations of our results at
\url{http://streetstyle.cs.cornell.edu}.

%% file: streetstyle/related.tex
\section{Related Work}\label{sec:streetstyle_related}
\noindent{\bf Visual discovery.} Researchers are beginning to mine
world-wide imagery to (1) discover visual trends and (2) measure
aspects of the world over space and time. Much of this work has looked
at {\em places}.  Doersch~\etal\ pioneered the idea of {\em
  computational geography} and explored the distinctive visual
characteristics of cities through Street View
imagery~\shortcite{doersch:siggraph:2012}, driven by weakly supervised
methods for finding discriminative
elements~\cite{singh:eccv:2012,doersch:nips:2013}.  Such visual
characteristics can also be correlated to other properties of cities,
such as perceived
safety~\cite{arietta:tvcg:2014,naik:cvprw:2014,dubey:eccv;2016}. These
correlations can then be used to automatically predict such properties
across a city simply from ground level images. Beyond analyses of
cities, computer vision techniques have been used to answer questions
such as ``is there snow in this photo?'' or ``are there clouds in this
webcam?'' which in turn can be used at scale to estimate maps of snow
or cloud cover over large regions of the
world~\cite{zhang:www:2012,murdock:iccv:2015}.

Other work has used large image collections to study people, including
explorations of how people move through
cities~\cite{crandall:www:2009}, the relationship between facial
appearance and
geolocation/time~\cite{islam:eurosip:2015,salem:wacv:2016}, analysis
of expressions and styles over a century in high school yearbook
photos~\cite{ginosar:iccvw:2015}, and tools for discovering visual
patterns that distinguish two
populations~\cite{matzen:iccv:2015}. Finally, Gebru \etal\ study
demographics across the US by detecting and analyzing cars (along with
their makes and models) in Street View
imagery~\shortcite{gebru:arxiv:2017}. Compared to this prior work, we
focus on a different domain, fashion and style, and apply our work to
a much broader sample of people by using worldwide images on social
media.

\smallskip \noindent{\bf Visual understanding of clothing.} Clothing
is a rich, complex visual domain from the standpoint of computer
vision, because clothing analysis combines a detailed attribute-level
understanding with social context.  We build on previous work that
describes people in terms of clothing and other fashion attributes,
such as ``wearing glasses,'' or ``short sleeves,'' and recognizes
these attributes in new
images~\cite{chen:eccv:2012,bourdev:iccv:2011,bossard:accv:2013,zhang:cvpr:2014}.
Beyond classifying attributes, other work also produces full
pixel-level clothing
segmentations~\cite{yamaguchi:cvpr:2012,yamaguchi:iccv:2013,yang:cvpr:2014},
or recognizes specific (or similar) products rather than general
attributes~\cite{di:cvprw:2013,vittayakorn:wacv:2015,kiapour:iccv:2015}.
Other work categorizes clothing in terms of explicit, named
styles. For instance, Kiapor~\etal\ used a game to analyze coarse
fashion styles such as ``hipster'' and ``goth'' to build inter-style
classifiers (i.e., is this photo goth or hipster?) and intra-style
ranking functions (i.e., how hipster is this person's
fashion?)~\shortcite{kiapour:eccv:2014}.  In our work, we train
classifiers using coarse-grained attributes, but then leverage this
training to organize images visually in order to perform more
fine-grained clustering.

\smallskip \noindent{\bf Clothing trends.} Spatial and temporal
fashion trends have been explored in computer vision, but usally on
small samples sizes and with limited statistical analysis.
Hidayati~\etal\ analyze catwalk images from NYC fashion shows to find
style trends in high-end fashion~\shortcite{hidayati:acmmm:2014}.
Simo-Serra~\etal\ analyzed \url{chictopia.com} to study correlations
between fashionability and other attributes such as wealth, as well as
temporal trends (e.g., a sudden spike in popularity of heels in
Manila)~\shortcite{simoserra:cvpr:2015}.  \cite{vittayakorn:wacv:2015}
showed that seasonal trends related to styles such as ``floral'',
``pastel'', and ``neon'' can be found using fashion classifiers, with
the peak occurring in the springtime.  \cite{he:www:2016} modeled
per-user fashion taste over time and found certain styles had a
resurgence in the late 2000s.  However, these trends were not
evaluated for statistical significance, and so it is challenging to
conclusively distinguish signal from noise. We argue that without the
massive amounts of data we advocate, it is difficult to establish such
significance.

\smallskip \noindent{\bf Clothing datasets.} Most previous clothing style
datasets have been very limited in scale, and biased towards images of the
fashion-conscious.  For example, several efforts have made use of
fashion-centric social networking sites such as \url{fashionista.com} and
\url{chictopia.com}~\cite{yang:cvpr:2014,simoserra:cvpr:2015,yamaguchi:bvmc:2015,yamaguchi:cvpr:2012,yamaguchi:iccv:2013}.
These sites enable users to upload photos and annotate articles of clothing for
the express purpose of modeling their style. Other work draws on online clothing
retailers such as Amazon, eBay, and ModShop to build similar
datasets~\cite{liu:acmmm:2012,di:cvprw:2013,kiapour:iccv:2015}.  Images from
these websites are relatively clean and organized according to tags,
facilitating the creation of annotated datasets on the scale of hundreds of
thousands of people~\cite{chen:eccv:2012,bossard:accv:2013}.  However, our goal
is to measure spatio-temporal trends over the entire world from real-world
images, and to obtain highly certain statistics, so very large data is
key---100K images is insufficient, once the data is sliced in time and space.
Furthermore, fashion sites are designed to engage the fashion-conscious, whereas
our goal is to analyze the world's populace at large.  Our approach is to build
a massive dataset from photos on social media, which can be gathered at larger
scale and are more representative of everyday fashions (hence the name {\it
StreetStyle}). However, these images are also much more noisy and are often not
tagged according to clothing.  Therefore, we annotate a small subset of our large
dataset manually, and use machine learning  to generalize the result to the rest
of the dataset.

%% file: streetstyle/data.tex
\section{Data}\label{sec:streetstyle_data} Out dataset consists of three key
parts: (1) photos, (2) the people in those photos, and (3) the
clothing attributes of those people.

\smallskip \noindent{\bf Photos.} The photos we use in our analysis
were acquired via Instagram, a popular mobile photography social
network.  We chose Instagram due to the sheer volume of uploads---an
advertised 95 million photos/videos per
day\footnote{\url{http://time.com/4375747/instagram-500-million-users/}}---as
well as their providing a public photo search
API.\footnote{\url{https://www.instagram.com/developer/}} Users can
query this API to retrieve images that have been uploaded within a 5
kilometer radius of a specified latitude and longitude and within 5
days of a specified date.

\begin{figure}[tp]
\begin{center}
\includegraphics[width=\columnwidth]{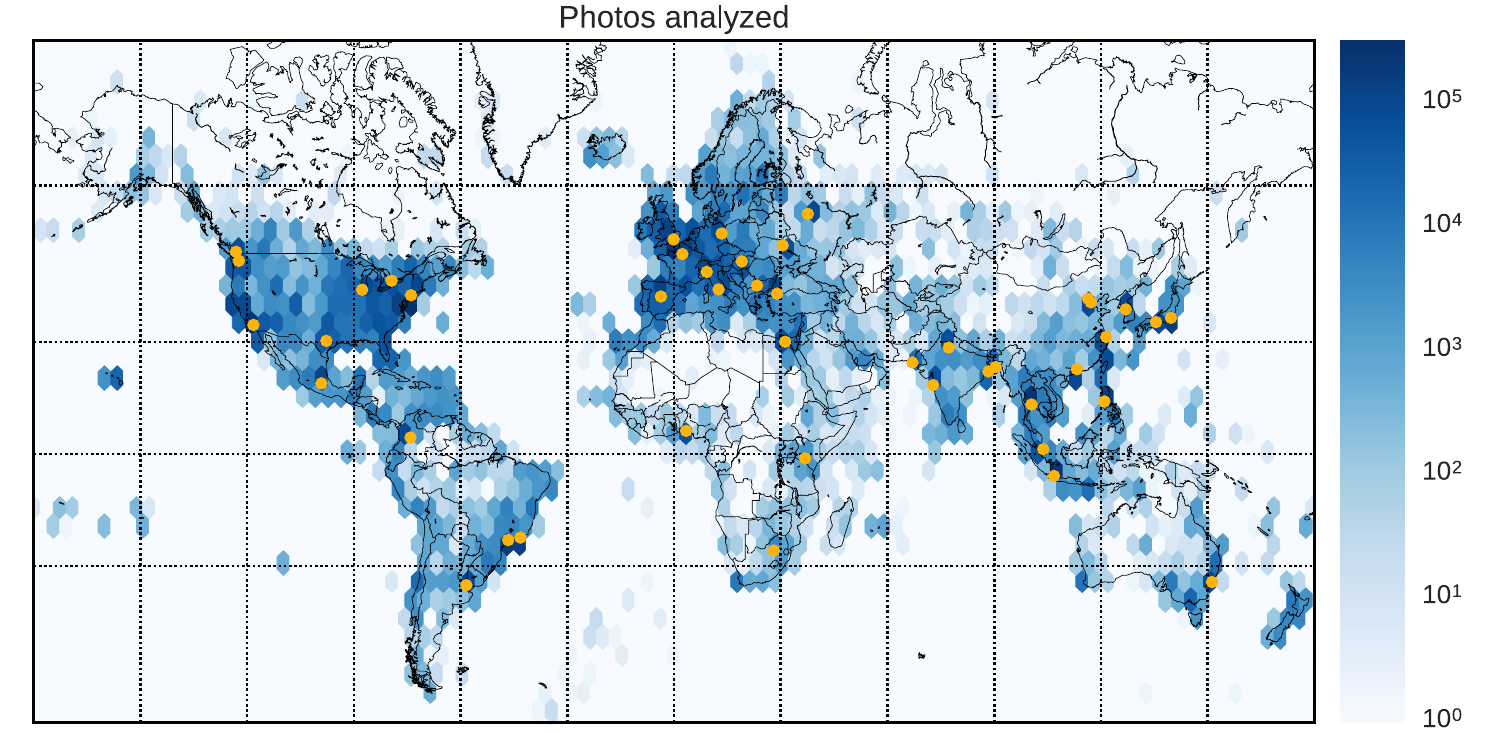}
\caption{{\bf Geo-spatial distribution of images collected from Instagram}
  (after filtering for images of people).  Images were collected
  according to two distributions: a set of the world cities (yellow
  dots), and a distribution over the entire Earth (blue). The shade of
  blue shows the number of photos downloaded from a particular region
  (according to a logarithmic
  scale).}\label{fig:streetstyle_distribution}
\end{center}
\end{figure}

\input{streetstyle/cities}

How should we sample photos geographically? To answer this question,
we consider two types of experiments we wish to conduct: (1) a
comparison of major world population centers,
and (2) a nation-, continent-, or world-wide analysis of styles and
trends.  To sample photos for goal (1) (analysis of urban centers), we
made a list of 44 major cities spanning six continents, each with a
lat/long approximating the city center, shown as yellow dots in
Figure~\ref{fig:streetstyle_distribution}. The complete list of cities
is shown in Table~\ref{tab:streetstyle_cities}.
For each city, we sampled photos centered on those coordinates from
Instagram.
To sample photos for goal (2) (globally distributed photos), we used
the Flickr 100M photo dataset~\cite{thomee:arxiv:2015} to derive a
distribution of photo uploads over the globe.  In particular, we
extracted the approximately 48 million geotags from Flickr 100M, and
used these geotags to compute a geographic distribution from which to
sample photos from Instagram.  This distribution (after filtering by
person detection, as discussed below) is shown in blue in
Figure~\ref{fig:streetstyle_distribution}.

For both sampling strategies, we uniformly sampled a 5-day window from
June 2013---the earliest date Instagram provided results---until June
2016.  For each photo, we record its geolocation and timestamp. In
total, we retrieved over 100 million photos.

\smallskip \noindent {\bf People.} To find people in photos, we ran
two out-of-the-box vision algorithms on each downloaded photo, one
to detect and localize faces, and the second to estimate the
visibility of the rest of the body. To detect faces, we used the API
provided by Face++.\footnote{\url{http://www.faceplusplus.com/}} The
output of the Face++ detector is a bounding box for the face, as well
as facial landmark keypoints.  We use the face bounding box to
geometrically align people in our analysis algorithms.  To determine
body visibility, we use the Deformable Part
Model~\cite{felzenszwalb:pami:2010,girshick:code:2012} and the
provided person detector trained on
VOC2010~\cite{everingham:site:voc2010}.

\begin{figure}[tp]
\begin{center}
\includegraphics[width=\columnwidth]{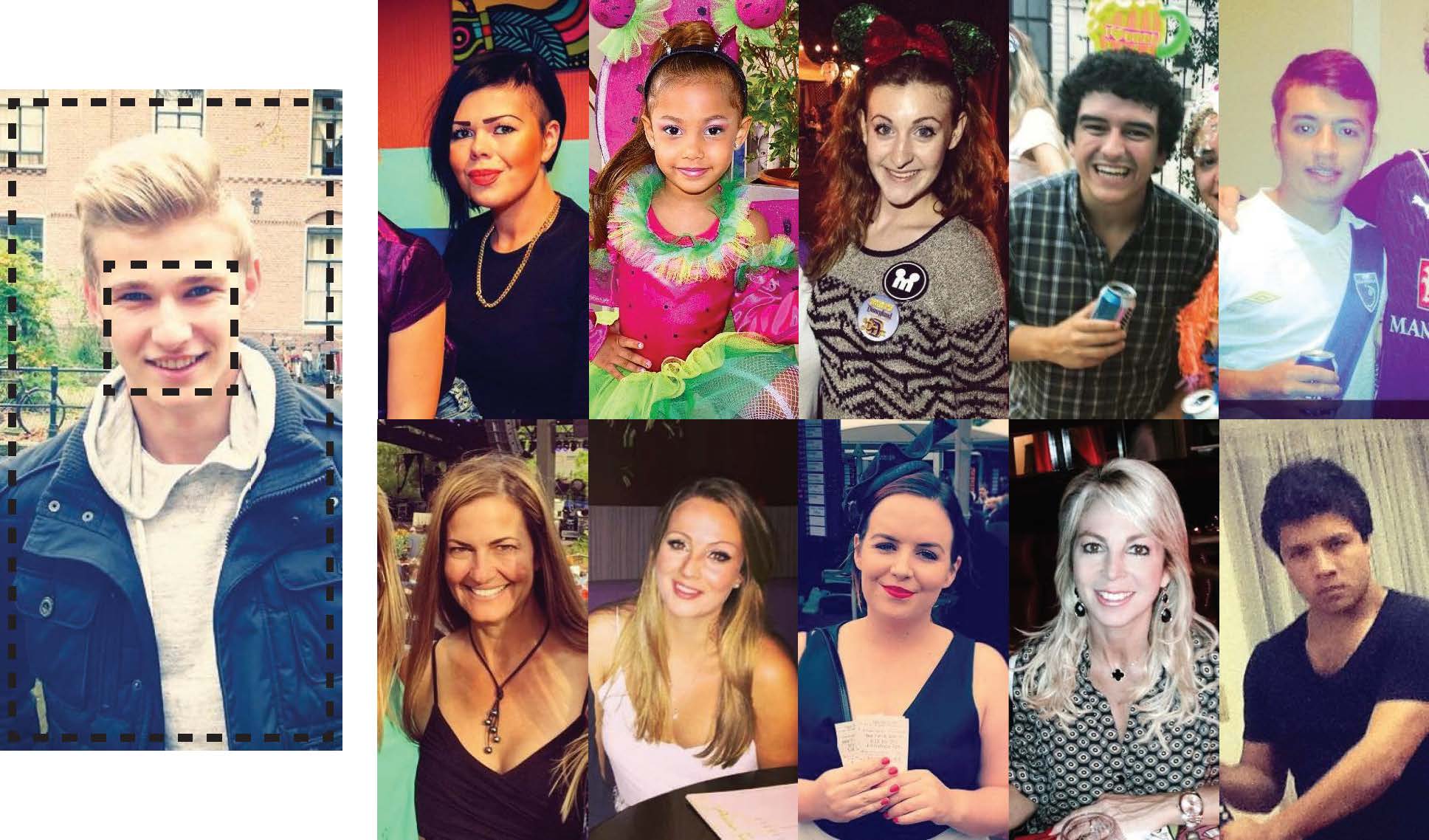}
\caption{Left: Given an approximate person detection and a precise
  face localization, we generate a crop with canonical position and
  scaling to capture the head and torso of the person.  Right: Example
  people from our dataset.}\label{fig:streetstyle_crop}
\end{center}
\end{figure}

Given a set of faces and visible bodies in a photo, we pair them up
using a simple distance-based heuristic.  For each face, we compute a
canonical image crop based on the position and scale of the detected
face, illustrated in Figure~\ref{fig:streetstyle_crop}.  If the crop
extends beyond the visible body bounding box, we discard the
detection.

In total, out of the more than 100 million photos we retrieved from
Instagram, 37.7 million had at least one successful person
detection. For our work, we chose to keep detections with at least the
face and torso visible.  We do not analyze the lower part of the body
since the legs are often occluded in online photos.  After generating
canonical crops from the face detections and filtering on face and
torso visibility, we gathered a total of 14.5 million images of
people.

\begin{table}
\begin{center}
\resizebox{\columnwidth}{!}{\begin{tabular}{|c|c|c|} \hline & {\bf No}
    & {\bf Yes} \\
\hline Wearing Jacket & 18078 & 7113 \\ Collar Presence & 16774 & 7299 \\
Wearing Scarf & 23979 & 1452\\ Wearing Necktie & 24843 & 827\\ Wearing Hat &
23279 & 2255\\ Wearing Glasses & 22058 & 3401\\ Multiple Layers & 15921 & 8829\\
&&\\ &&\\ &&\\ &&\\ &&\\ &&\\ &&\\ \hline
\end{tabular}
{\hskip 0.1cm}
\begin{tabular}{|c|}
\hline {\bf Major} \\ {\bf Color} \\ \hline Black (6545) \\ White (4461) \\ 2+ colors (2439)
\\ Blue (2419) \\ Gray (1345)\\ Red (1131)\\ Pink (649)\\ Green (526)\\ Yellow
(441)\\ Brown (386)\\ Purple (170)\\ Orange (162)\\ Cyan (33)\\ \hline
\end{tabular}
\begin{tabular}{|c|}
\hline {\bf Clothing} \\ {\bf Category} \\ \hline Shirt (4666) \\ Outerwear (4580) \\
T-shirt (4580) \\ Dress (2558) \\ Tank top (1348) \\ Suit (1143) \\ Sweater
(874) \\ \hline \hline {\bf Sleeve} \\ {\bf Length} \\ \hline Long sleeve (13410) \\ Short
sleeve (7145) \\ No sleeve (3520) \\ \\ \hline
\end{tabular}
\begin{tabular}{|c|}
\hline {\bf Neckline} \\ {\bf Shape} \\ \hline Round (9799) \\ Folded (8119) \\ V-shape
(2017) \\ \hline \hline {\bf Clothing} \\ {\bf Pattern} \\ \hline Solid (15933) \\ Graphics
(3832) \\ Striped (1069) \\ Floral (885) \\ Plaid (532) \\ Spotted (241)\\ \\ \\
\hline
\end{tabular}
}
\caption{{\bf Fashion attributes in our dataset}, along with the number of
  MTurk annotations collected for each attribute.  Left: Binary
  attributes.  Right: Attributes with three or more
  values.}\label{table:streetstyle_attributes}
\end{center}
\end{table}

\smallskip \noindent{\bf Clothing annotations.} For each person in our
corpus of images, we want to extract fashion information. To do so, we
first collect annotations related to fashion and style for a subset of
the data, which we later use to learn attribute classifiers that can
be applied to the entire corpus. Example attributes include: {\em What
  kind of pattern appears on this shirt?  What is the neckline? What
  is the pattern?  Is this person wearing a hat?}  We created a list
of several clothing attributes we wished to study a priori, such as
clothing type, pattern, and color.  We also surveyed prior work on
clothing attribute recognition and created a consolidated list of
attributes that we observed to be frequent and identifiable in our
imagery.  We began with several of the attributes presented by Chen
\etal~\cite{chen:eccv:2012} and made a few adjustments to more clearly
articulate neckline type as in the work of Di
\etal~\cite{di:cvprw:2013}. Finally, we added a few attributes of our
own, including \emph{wearing-hat} and \emph{wearing-jacket}, as well
as an attribute indicating whether or not more than one layer of
clothing is visible (e.g., a suit jacket over a shirt or an open
jacket showing a t-shirt
underneath). Table~\ref{table:streetstyle_attributes} shows our full
list of attributes.

We annotated a 27K-person subset of our person image corpus with these
attributes using Amazon Mechanical Turk.  Each Mechanical Turk user
was given a list of 40 photos, presented one-by-one, and asked to
select one of several predefined attribute labels.  In addition, the
user could indicate that the person detection was faulty in one of
several ways, such as containing more than one person or containing no
person at all.  Each set of 40 photos contained two photos with known
ground truth label as sentinels.  If a worker failed at least five
sentinels and their sentinel failure rate was above 20\percent, we
prevented them from performing any more work for our tasks.
Compensation for a single attribute on a set of 40 photos was 0.10
USD.  Average wage was approximately 4 USD / hour.  An attribute for
an image was labeled by five separate workers, and a different set of
five workers labeled each attribute for an image.  A label was
accepted if 3 / 5 workers agreed, otherwise it was excluded from the
data.  In total these annotations cost approximately 4K USD.  We refer
to our annotated dataset as \StreetStyle.

\smallskip \noindent{\bf Dataset bias and limitations.} Any dataset
has inherent bias, and it is important to be cognizant of these biases
when drawing conclusions from the data.  Instagram is a mobile social
network, so participants are more likely to be in regions where
broadband mobile Internet is available, and where access to Instagram
is not censored or blocked.  In some areas, people might be more
likely to upload to a competitor service than to Instagram.  People of
certain ages might be more or less likely to upload to any social
network.  Cultural norms can affect who, when, where, and what people
photograph.  The face detector we use is an off-the-shelf component
for which no reported statistics regarding bias for age, gender, and
race are made available.  The person detector has also not been
evaluated to determine bias in these factors either.  These factors
impact who will and will not be properly added to our dataset, which
could introduce bias. As vision methods mature, analyses for such bias
will become increasingly important.

%% file: streetstyle/cities.tex
\begin{table}[tbh]
\begin{center}
\begin{tabular}{|cccc|}
\hline Austin & Bangkok & Beijing & Berlin \\ Bogotá & Budapest & Buenos
Aires & Cairo\\ Chicago & Delhi & Dhaka & Guangzhou \\ Istanbul & Jakarta &
Johannesburg & Karachi \\ Kiev & Kolkata & Lagos & London \\ Los Angeles &
Madrid & Manila & Mexico City \\ Milan & Moscow & Mumbai & Nairobi\\ New
York City & Osaka & Paris & Rio de Janeiro\\ Rome & São Paulo & Seattle &
Seoul \\ Shanghai & Singapore & Sofia & Sydney \\ Tianjin & Tokyo & Toronto &
Vancouver\\ \hline
\end{tabular}
\caption{Cities sampled for our analysis.}\label{tab:streetstyle_cities}
\end{center}
\end{table}

%% file: streetstyle/method.tex
\section{Machine Learning Methodology}\label{sec:streetstyle_method}

To make measurements across the millions of photos in our dataset, we
wish to generalize labeled data from \StreetStyle to the entire
corpus. For instance, to measure instances of glasses across the
world, we want to use all of the {\em wearing-glasses} annotations as
training data and predict the {\em wearing-glasses} attribute in
millions of unlabeled images.  To do so we use Convolutional Neural
Networks (CNNs)~\cite{krizhevsky:nips:2012}. A CNN maps an input
(e.g., the pixels of an image), to an output (e.g., a binary label
representing whether or not the person is wearing glasses), by
automatically learning a rich hierarchy of internal image features
given training data.  In this section, we describe the architecture of
our particular CNN, how we trained it using our labeled attribute
dataset, and how we evaluated the classifier's performance on the
clothing attribute prediction task.  We then discuss how and when it
is appropriate to generate predictions from this CNN across millions
of photos and use these predictions to estimate real-world fashion
statistics.

\subsection{Learning attributes} A key design decision in creating a CNN is
defining its network architecture, or how the various types of
internal image processing layers are composed. The penultimate layer
of a typical CNN outputs a multi-dimensional (e.g., 1024-D) feature
vector, which is followed by a ``linear layer'' (matrix product)
operation that outputs one or more prediction scores (e.g., values
indicating the probability of a set of attributes of interest, such as
{\em wearing-glasses}).  The base architecture for our CNN is the
``GoogLeNet'' architecture~\cite{szegedy:cvpr:2015}.  While several
excellent alternatives exist (such as VGG~\cite{simonyan:arxiv:2014}),
GoogLeNet offers a good tradeoff between accuracy on tasks such as
image classification~\cite{russakovsky:ijcv:2015} and speed.  As with
other moderate-sized datasets (fewer than 1 million training
examples), it is difficult to train a CNN on our attribute prediction
task from scratch without overfitting.  Instead, we start with a CNN
pretrained on image classification (on ImageNet ILSVRC2012) and
fine-tune the CNN on our data.  After ImageNet training, we discard
the last linear layer of the network to expose the 1024-dimensional
feature output and then append several $1024\times N_i$ linear layers
in parallel where $N_i$ is the number of class labels for attribute
$i$ (for instance, $N_i = 2$ for a binary attribute such as {\em
  wearing-glasses}, and $N_i = 3$ for neckline shape as we consider
three possible shapes).

\smallskip \noindent{\bf Training details.} We train the CNN as
follows. For each attribute, we use stratified sampling to first
select an attribute label and then a particular example image with
that label.  This sampling strategy counteracts the implicit imbalance
within each attribute (e.g., wearing a hat is much less common than
not wearing a hat).  We do this 32 times to build a single mini-batch
of images.  Using this mini-batch, we apply the forward and backward
passes of the CNN to compute the parameter gradient with respect to
the cross-entropy loss. We do this for each attribute, accumulating
the gradients.  Then we use stochastic gradient descent with momentum
$= 0.9$, learning rate $= 10^{-2}$, and weight decay $= 10^{-4}$ to
update the parameters of the CNN.  We fine-tune the CNN for 6,000
iterations ($\times 12$ attributes) after which the mean class
accuracy on a validation set stopped increasing.  80\percent\ of
\StreetStyle\ was used for training, 10\percent\ was used as
validation to determine when to stop training as well as for the
additional analysis in this section, and 10\percent\ was used as a
test set for a final evaluation of the attribute classification task.

\begin{table}[t]
\begin{center}
\resizebox{\columnwidth}{!}{\begin{tabular}{|c|c|c|c|} \hline {\bf
      Attribute} &
\begin{tabular}{@{}c@{}}{\bf Our CNN} \\(acc. / mean class. acc.)\end{tabular} &
{\bf Random guessing}& {\bf Majority guessing} \\ \hline \hline wearing jacket & 0.869 /
0.848  &0.5 / 0.5 & 0.698 / 0.5 \\ clothing category & 0.661 / 0.627 &0.142 /
0.142 & 0.241 / 0.142 \\ sleeve length &0.794 / 0.788 & 0.333 / 0.333 & 0.573 /
0.333 \\ neckline shape & 0.831 / 0.766 &0.333 / 0.333 & 0.481 / 0.333 \\ collar
presence & 0.869 / 0.868 & 0.5 / 0.5 & 0.701 / 0.5 \\ wearing scarf & 0.944 /
0.772 & 0.5 / 0.5 & 0.926 / 0.5 \\ wearing necktie & 0.979 / 0.826 & 0.5 / 0.5 &
0.956 / 0.5\\ clothing pattern & 0.853 / 0.772 & 0.166 / 0.166 & 0.717 / 0.166
\\ major color & 0.688 / 0.568 & 0.077 / 0.077 & 0.197 / 0.077 \\ wearing hat &
0.959 / 0.917 & 0.5 / 0.5 & 0.904 / 0.5\\ wearing glasses & 0.982 / 0.945 & 0.5
/ 0.5 & 0.863 / 0.5 \\ multiple layers & 0.830 / 0.823 & 0.5 / 0.5 & 0.619 /
0.5\\ \hline
\end{tabular}}
\caption{{\bf Test set performance for the trained CNN.}  Since each
  attribute has an unbalanced set of test examples, we report both
  accuracy and mean classification accuracy.  For reference, we also
  include baseline performance scores for random guessing and majority
  class guessing.}\label{table:streetstyle_accuracy}
\end{center}
\end{table}

\begin{figure}[t]
\begin{center}
\includegraphics[width=\columnwidth]{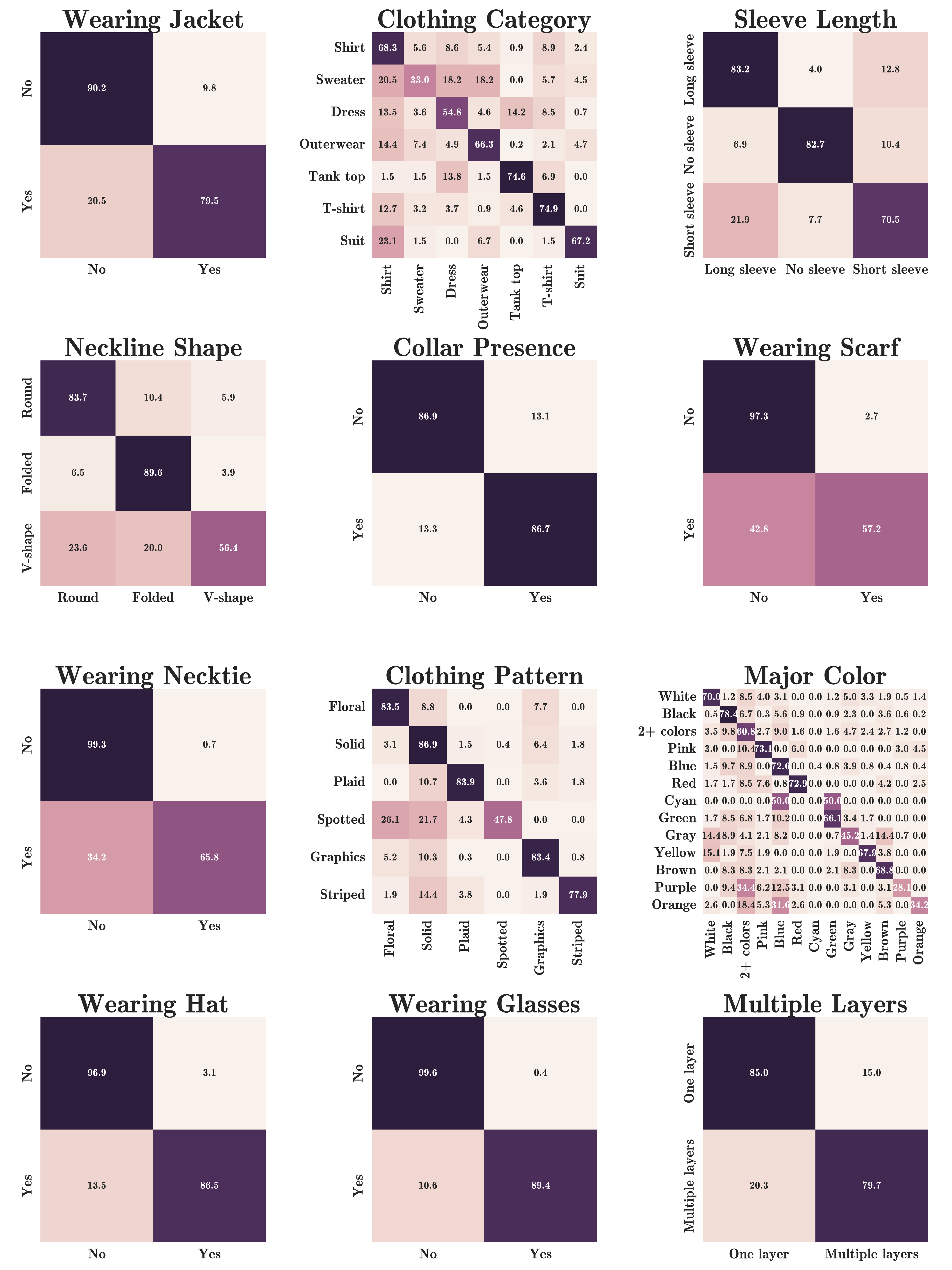}
\caption{{\bf Confusion matrices for each clothing attribute.} Given class
  labels are vertical and predicted class labels are horizontal.  Each
  row has been normalized such that a cell indicates the probability
  of a classification given some true
  label.\label{fig:streetstyle_confusion}}
\end{center}
\end{figure}

Table~\ref{table:streetstyle_accuracy} summarizes the accuracy and
mean class accuracy of our learned classifiers for each attribute on
the held-out test set.  Figure~\ref{fig:streetstyle_confusion} shows a
set of confusion matrices, one per attribute, illustrating which
attributes tend to get confused for one another (e.g., spotted shirts
can be mistaken for floral shirts).  While the classifiers perform
quite well in general (e.g., {\em wearing-hat} has an accuracy over
90\percent), even state-of-the-art CNNs make mistakes, and so all of
the classifiers have an inherent error level.  We now discuss how we
take this error into account when using these classifiers.

\subsection{Measuring attributes at scale} Our aim is not just to build a
classifier for each clothing attribute and maximize some performance
criterion on a test set, but to take that learned classifier, apply it
to a much larger corpus of images spanning years worth of the world's
photos, and provide a tool for discovering interesting trends.
However, the classifiers are not perfect, and any real-world
measurement is going to have some noise.  While it is inevitable that
a classifier will have some non-zero variance, the hope is that the
estimation is unbiased, that is, as we increase the sample size, a
computed statistic, such as the detected percentage of people wearing
hats in photos, will converge on the true percentage of people wearing
hats.  However, by examining the confusion matrices in
Figure~\ref{fig:streetstyle_confusion} it becomes clear that this will
not be the case.  For example, the proportion of images featuring
people wearing neckties is much less than 50\percent.  For this
attribute, the classifier predicts \textbf{No} correctly
99\percent\ of the time, but \textbf{Yes} correctly only
65\percent\ of the time.  Therefore, the estimate of the percentage of
people wearing neckties in photos will be biased towards \textbf{No}.

\begin{figure*}[h]
\begin{center}
\includegraphics[width=\textwidth]{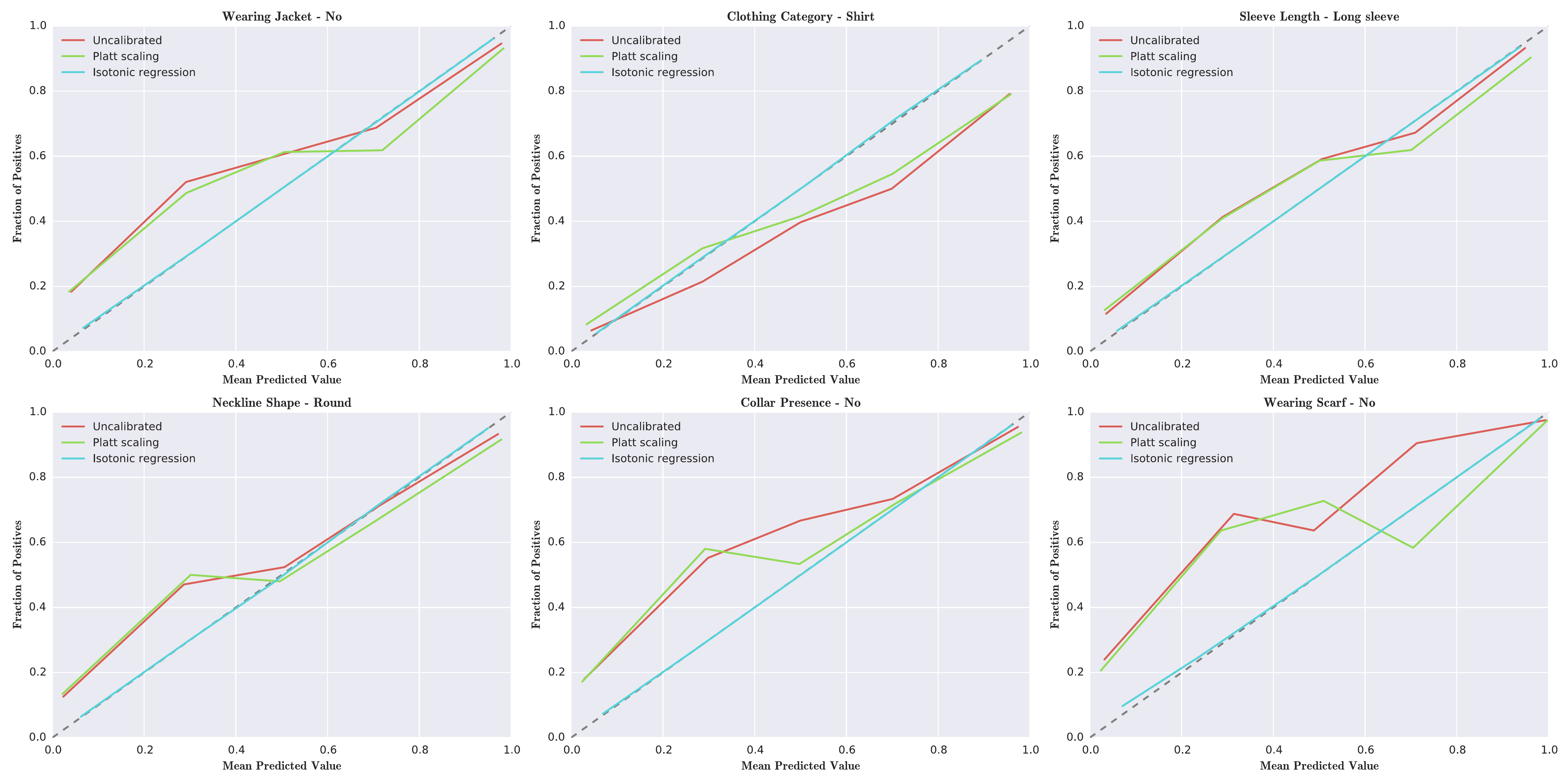}
\caption{{\bf Reliability curves for several clothing attributes.}  A
  well-calibrated reliability curve has fraction of positives equal to
  mean predicted value.  Neural networks tend to produce
  well-calibrated probabilities, but significant improvement is made
  using Isotonic regression for additional
  calibration.\label{fig:streetstyle_reliability}}
\end{center}
\end{figure*}

One way to mitigate this bias is to calibrate the scores of a
classifier using a function that models the posterior probability of
the classification being correct given the classification score.
There are several methods for this type of calibration, including
Platt scaling (an application of logistic
regression)~\cite{platt:book:1999} and isotonic regression. Others
have noted that neural networks tend not to require calibration as
severely as classifiers such as SVMs and boosted decision
trees~\cite{niculescu:icml:2005}.  Nevertheless, we found that our
networks benefited from calibration by applying isotonic regression to
our validation set.  Figure~\ref{fig:streetstyle_reliability} shows
the generalization performance of calibration for several attribute
labels.  Note that after isotonic regression, the curves are very
close to the identify function ($y=x$), which means that these curves
are properly calibrated.
In our experiments, half of the validation set was used for training the
regressor and the other half was used to generate reliability curves.
Isotonic regression typically requires more data to avoid overfitting,
yet is more flexible when the calibration required is not sigmoidal,
an underlying assumption of Platt scaling.  The generalization shows
that these estimates do not exhibit severe overfitting and therefore
we opt to use isotonic regression trained on the entire validation set
to calibrate the scores.

\subsection{Visually consistent style clusters} \label{subsec:clusters} 

\begin{figure}[tp]
\begin{center}
\includegraphics[width=\columnwidth]{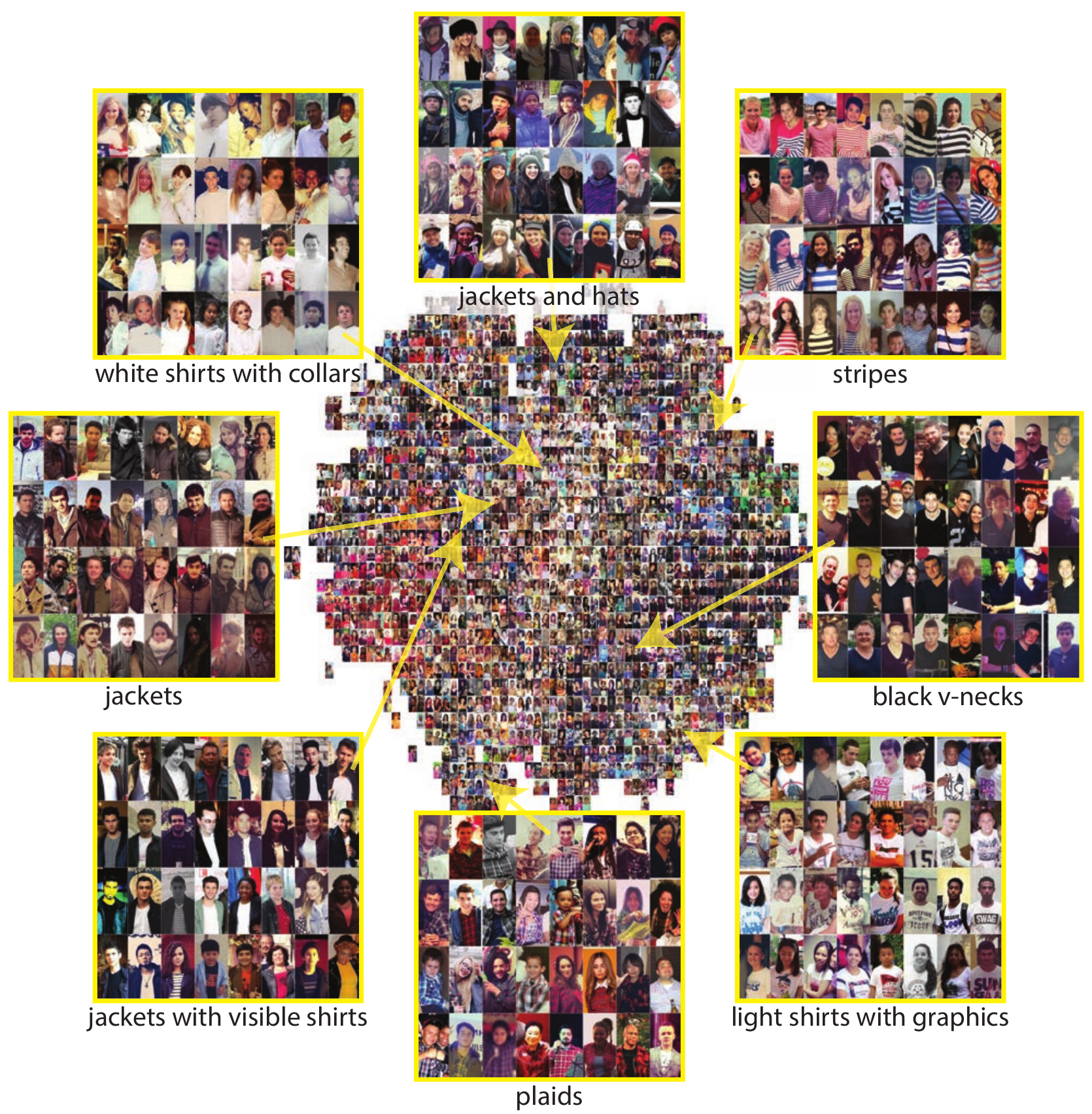}
\caption{{\bf t-SNE visualization of our learned fashion feature
    space.}  The style embedding is 1024-dimensional and is visualized
  in 2D here.  Several combinations of attributes are highlighted.
  For example, white shirts with text, plaid shirts, black v-neck
  shirts, etc., are styles discoverable in this embedding. Zoom to see
  detail.} \label{fig:tsne}
\end{center}
\end{figure}

\begin{figure*}[tp]
\begin{center}
\includegraphics[width=\textwidth]{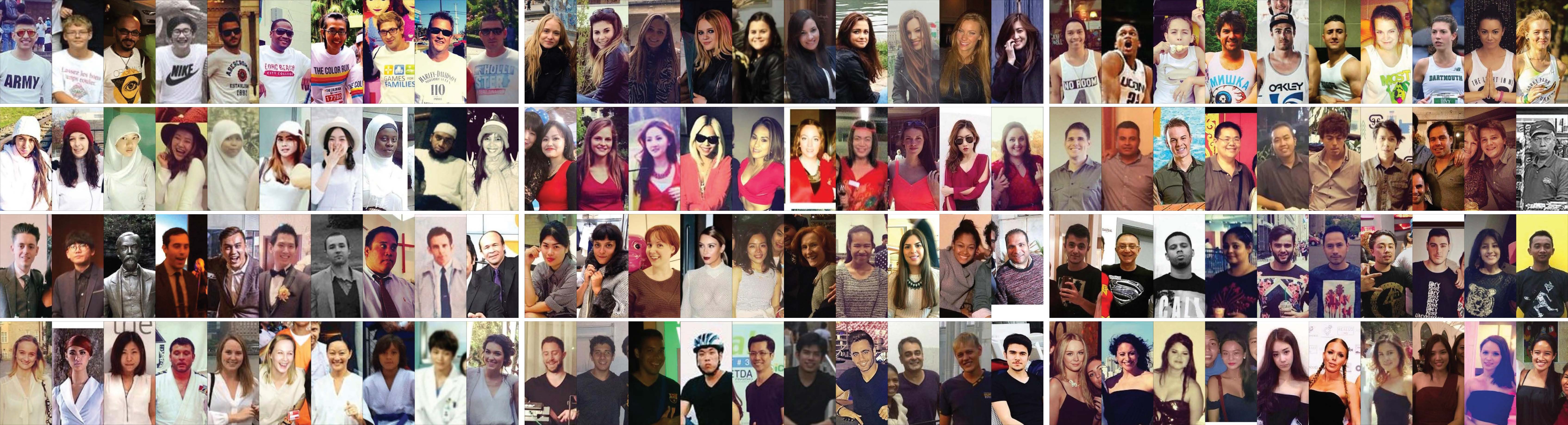}
\caption{{\bf Visualization of 12 style clusters}, showing images
  closest to the centroid of each cluster.  Clusters are shown in no
  particular order. Each style cluster reveals a visual theme as a
  combination of attributes, such as ``white graphic tees and
  glasses'' (top-left cluster). \label{fig:streetstyle_gmm_clusters}}
\end{center}
\end{figure*}

The predicted attributes are immediately useful in plotting and
analyzing trends, such as ``how is the proportion of people wearing
black in Los Angeles changing over time?'' as we demonstrate in
Section~\ref{sec:streetstyle_analysis}.  However, we would like to
analyze clothing styles beyond these attributes in a number of ways:
\begin{packed_item}
\item Identify common, visually correlated combinations of these basic
  attributes (e.g., {\em blue sweater with jacket and wool hat}).
\item Identify styles that appear more frequently in one city versus
  another or more frequently during particular periods of time.
\item Identify finer-grained, visually coherent versions of these
  elements (e.g., sports jerseys in a particular style).
\end{packed_item}

To achieve these goals, we use clustering to identify recurring {\em
  visual themes} in the embedded space of people. Recall that the
penultimate layer of our network is a 1024-dimensional feature space
where distinct fashion attributes are linearly
separable. Figure~\ref{fig:tsne} shows a visualization of this feature
space projected to 2D using t-SNE~\cite{maaten:jmlr:2008}. Within this
1024-D feature space, people images are organized according to visual
fashion attributes, through the mechanism of training on our
\StreetStyle dataset.  The ideas is that by clustering images in this
1024-D embedded feature space, we can discover and reveal visual
themes such as those described above.  We refer to such clusters as
{\bf style clusters}.
Once we identify style clusters, we can further characterize cities
and times in terms of these clusters---for instance, we might discover
that some clusters are heavily correlated with one or two particular
cities,
as we explore in Section~\ref{sec:streetstyle_analysis}.
Figure~\ref{fig:streetstyle_gmm_clusters} shows an example of several
style clusters found by our approach.

To find style clusters, we ran a clustering algorithm on a subset of
the full dataset, for efficiency and to achieve balance between
different times and regions of the world. In particular, we divided
our images into bins by city and by week (e.g., Paris in week 26 of
2015). For bins with fewer than $N$ images, we selected all images,
and for bins with more than $N$ images, we randomly selected $N$
images. For our experiments, we set $N = 4000$, for a total of 5.4M
sample images for clustering in total.

For each cropped person image in this set, we compute its 1024-D CNN
feature vector and $L_2$ normalize this vector. We run PCA on these
normalized vectors, and project onto the top principal components that
retain 90\percent\ of the variance in the vectors (in our case, 165
dimensions). To cluster these vectors, we used a Gaussian mixture
model (GMM) of 400 components with diagonal covariance matrices.  Each
person is assigned to the mixture component (style cluster) which
maximizes the posterior probability.  The people assigned to a cluster
are then sorted by their Euclidean distance from the cluster center,
as depicted in Figure~\ref{fig:streetstyle_gmm_clusters}.  Although
more clusters would increase the likelihood of the data under the
model, it would also lead to smaller clusters, and so we selected 400
as a compromise between retaining large clusters and maximizing the
likelihood of the data.

The resulting 400 style clusters have sizes ranging from 1K to
115K. Each style cluster tends to represent some combination of our
chosen attributes, and together form a global visual vocabulary for
fashion. We assign further meaning to these clusters by ranking them
and ranking the people within each cluster, as we demonstrate in
Section~\ref{sec:vis_styles}, with many such ranked clusters shown in
Figures~\ref{fig:streetstyle_rank_cities} and
\ref{fig:sorted_clusters}.

%% file: streetstyle/analysis.tex
\section{Exploratory Analysis}\label{sec:streetstyle_analysis}
A key goal of our work is to take automatic predictions from machine
learning and derive statistics from which we can find trends.  Some of
these may be commonsense trends that validate the approach, while
others may be unexpected. This section describes several ways to
explore our fashion data, and presents insights one can derive through
this exploratory data analysis.

\subsection{Analyzing trends}

\begin{figure}[tp]
\begin{center}
\includegraphics[width=\columnwidth]{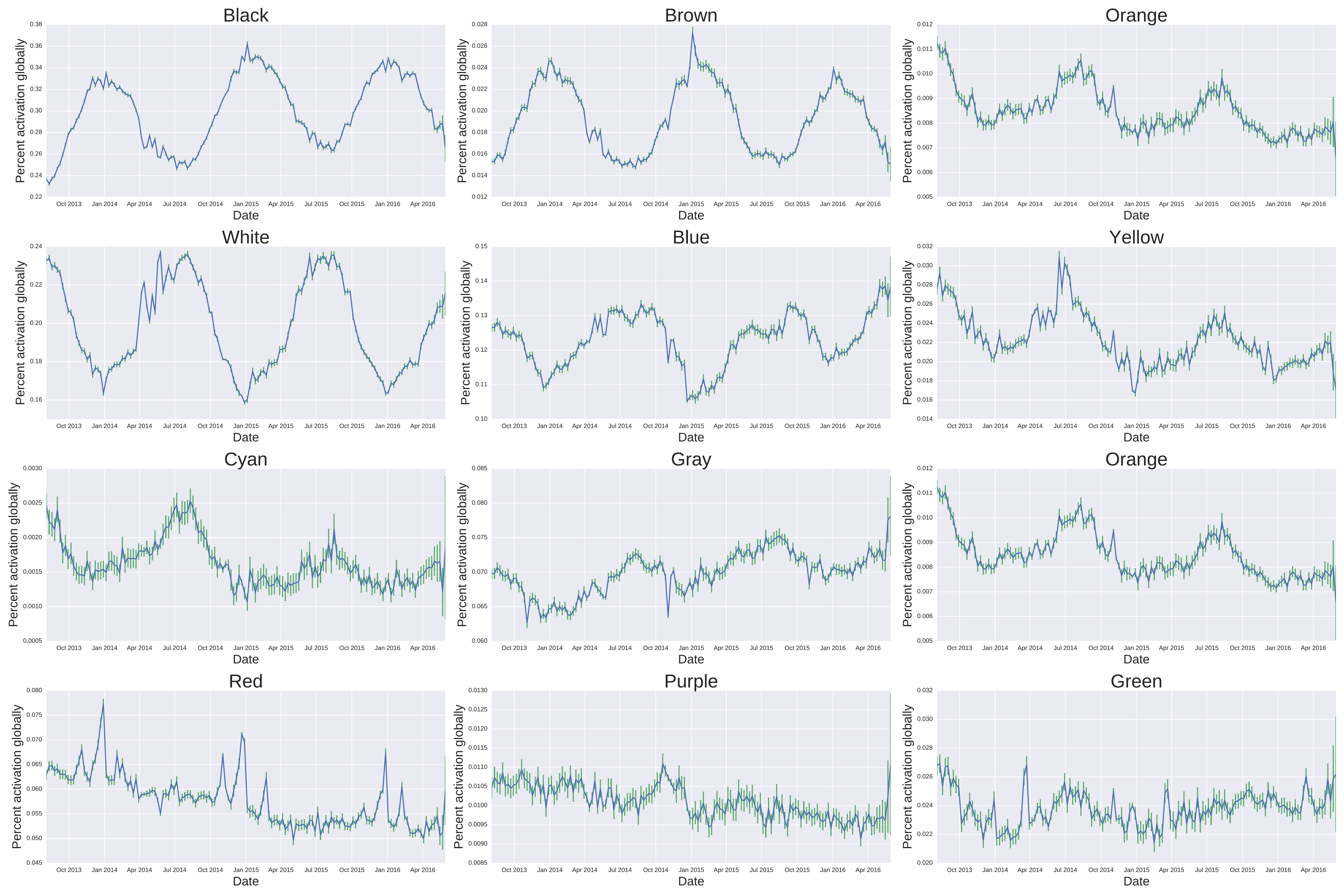}
\caption{{\bf Mean major color score for several different colors.}  The means are
computed by aggregating measurements per 1 week intervals and error bars
indicate 95\percent\ confidence interval.  Colors such as brown and black are
popular in the winter whereas colors such as blue and white are popular in the
summer.  Colors such as gray are trending up year over year whereas red, orange
and purple are trending down.\label{fig:streetstyle_color}}
\end{center}
\end{figure}

\begin{figure}[t]
\begin{center}
\includegraphics[width=\columnwidth]{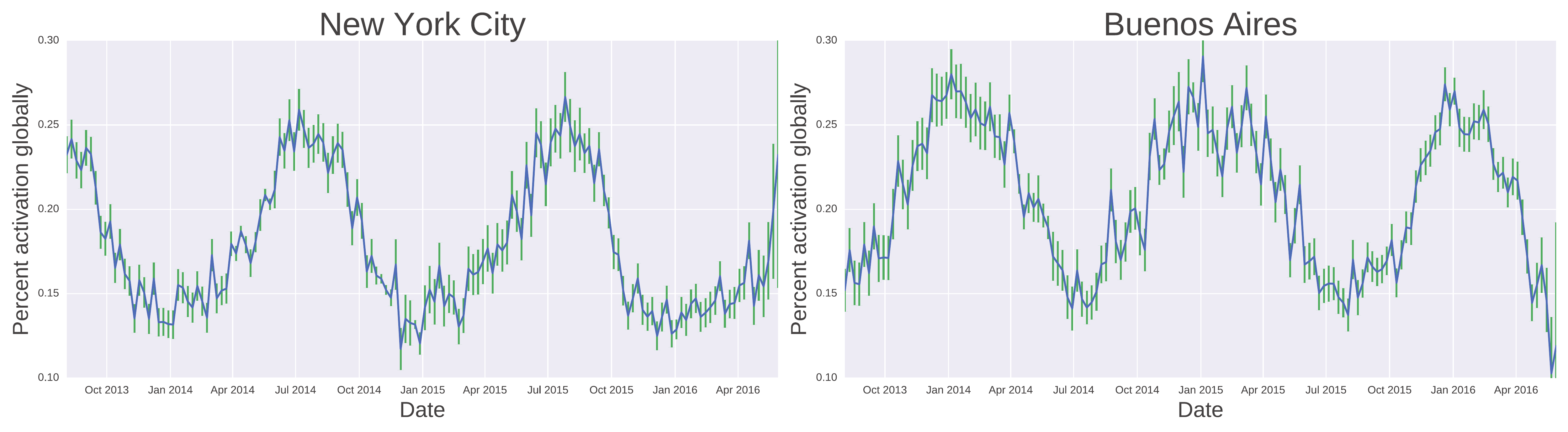}\\
\includegraphics[width=\columnwidth]{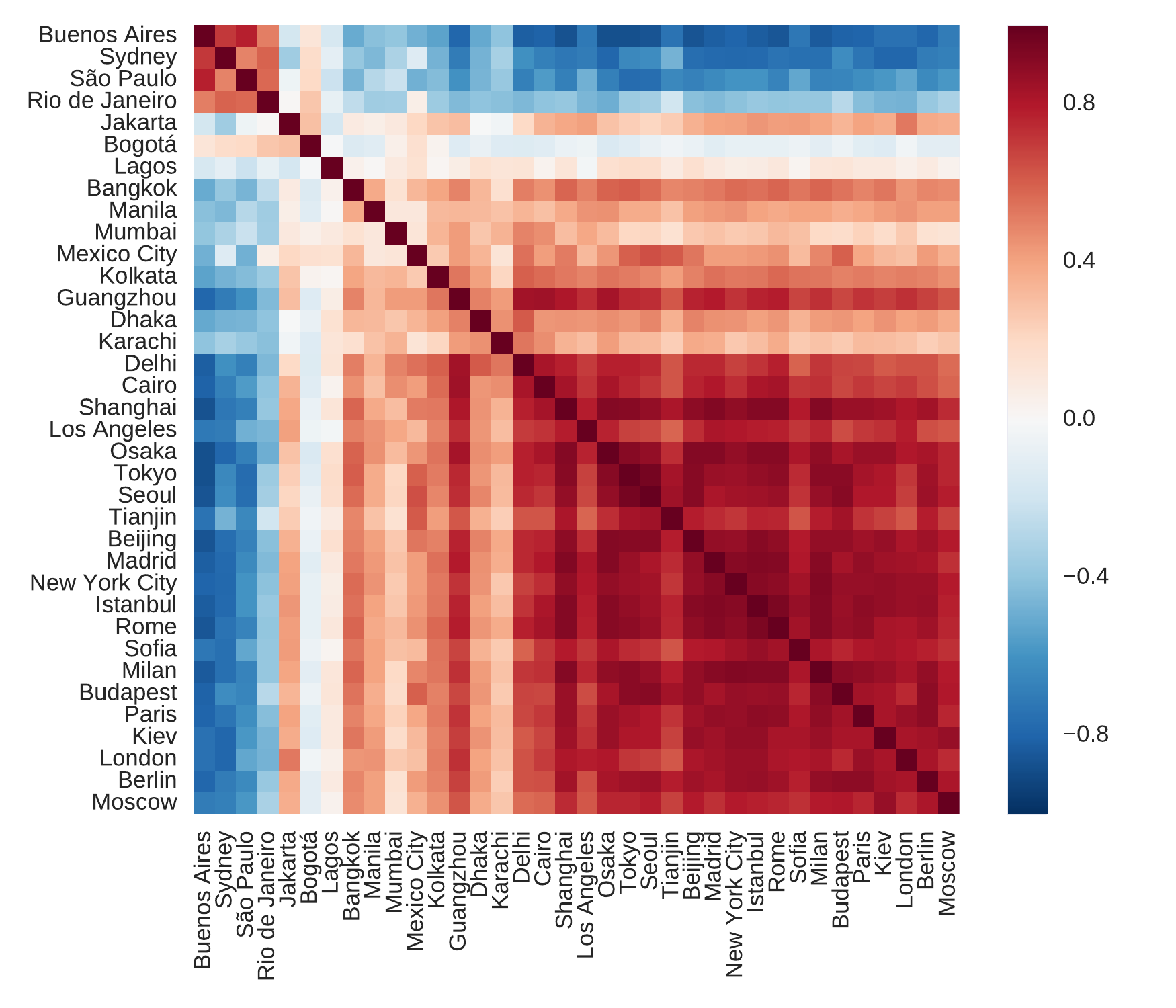}
\caption{Top: Mean major color frequency for the color white for two
  cities highlighting flipped behavior between the Northern and
  Southern Hemisphere.  Means are computed by aggregating measurements
  per 1 week intervals and error bars indicate a
  95\percent\ confidence interval.  Bottom: Pearson's correlation
  coefficients for the mean prediction score of the color white across
  all pairs of major world cities.  Cities have been ordered by
  latitude to highlight the relationship between the northern and
  southern hemispheres and their relationship to the seasonal trend of
  the color white.}\label{fig:streetstyle_color_cities}
\end{center}
\end{figure}

\begin{figure}[t]
\begin{center}
\includegraphics[width=\columnwidth]{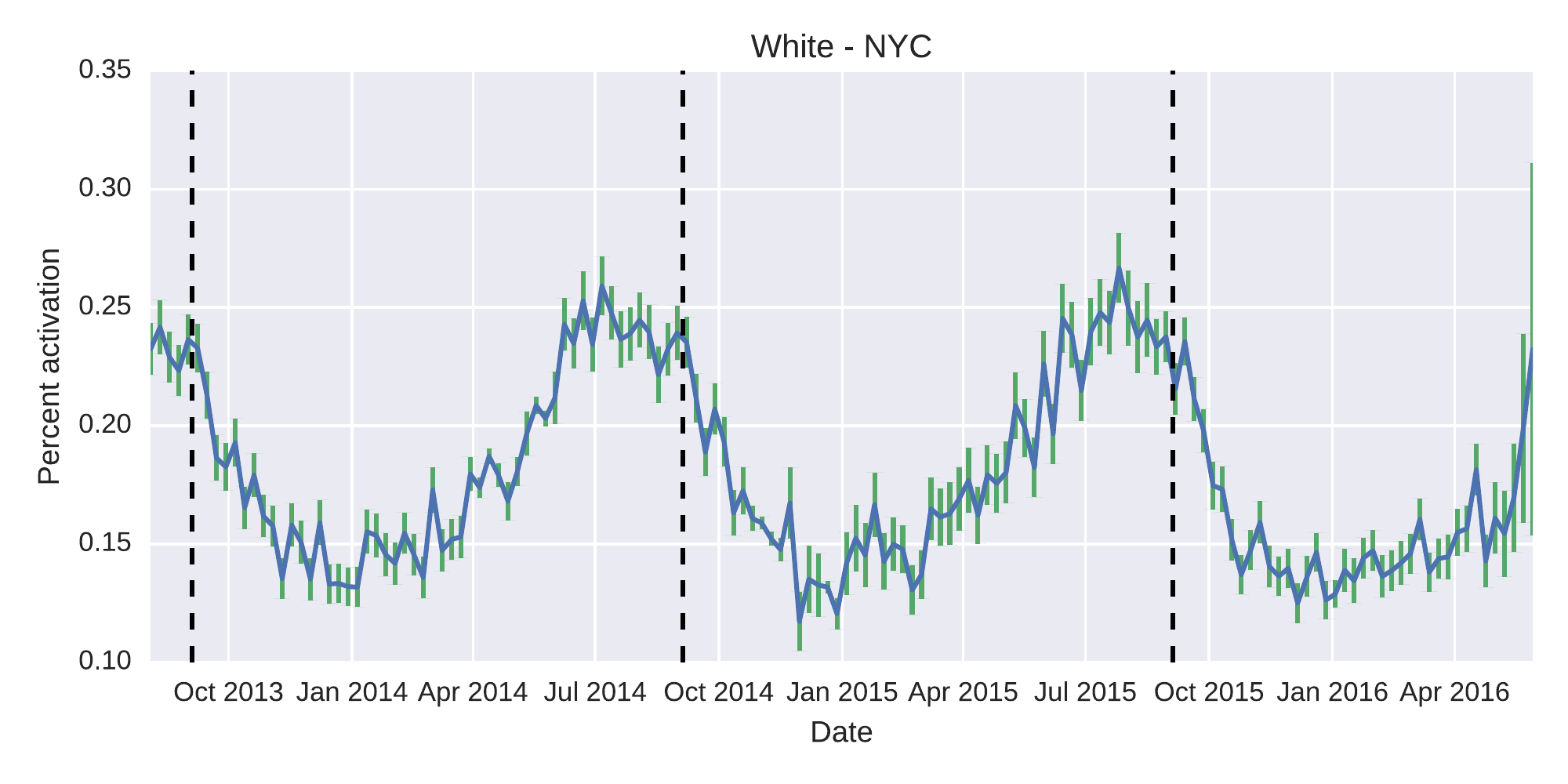}
\caption{{\bf Mean frequency for the color white in NYC.}  Means are
  computed by aggregating measurements per 1 week intervals and error
  bars indicate a 95\percent\ confidence interval.  Dashed lines mark
  Labor Day.}\label{fig:streetstyle_labor_day}
\end{center}
\end{figure}

\noindent{\bf Color.} How does color vary over time?
Figure~\ref{fig:streetstyle_color} shows plots of the frequency of
appearance (across the entire world) of several colors over time.
White and black clothing, for example, exhibit a highly periodic
trend; white is at its maximum ($>20\percent$ frequency) in September,
whereas black is nearly reversed, much more common in January.
However, if we break down these plots per city, we find that cities in
the Southern Hemisphere flip this pattern, suggesting a correlation
between season and clothing color.
Figure~\ref{fig:streetstyle_color_cities} shows such plots for two
cities, along with a visualization of all pairs of cities correlated
by their white color temporal profiles, with cities reordered by
latitude to highlight this seasonal effect. Cities in the same
hemisphere tend to be highly correlated with one another, while cities
in opposite hemispheres are highly negatively correlated.  In the
United States, there is an oft-cited rule that one should not wear
white after the Labor Day holiday (early September). Does this rule
comport with actual observed behavior? Using our method, we can
extract meaningful data that supports the claim that there is a
significant decrease in white clothing that begins mid-September,
shortly after Labor Day, shown in
Figure~\ref{fig:streetstyle_labor_day}.

\begin{figure}[t]
\begin{center}
\includegraphics[width=\columnwidth]{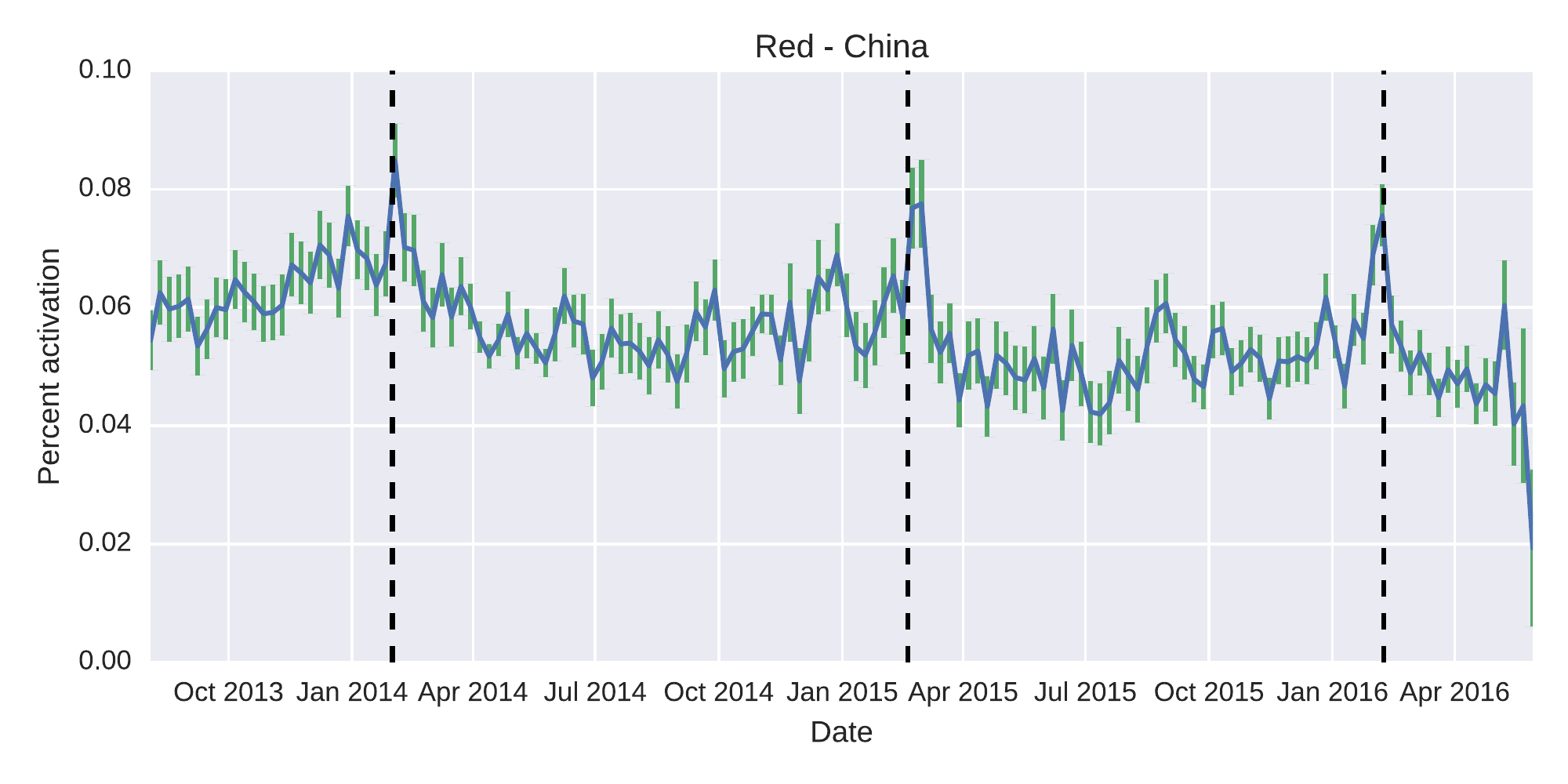}
\caption{{\bf Mean frequency for the color red in China.}  Means are
  computed by aggregating measurements per 1 week intervals and error
  bars indicate a 95\percent\ confidence interval.  Dashed lines mark
  Chinese New Year.}\label{fig:streetstyle_chinese_new_year}
\end{center}
\end{figure}

Other colors exhibit interesting, but more subtle trends.  The color
red is much less periodic, and appears to have been experiencing a
downward trend, down approximately 1\percent\ since a few years
ago. There are also several spikes in frequency that appear each year,
including one small spike near the end of each October and a much
larger spike near the end of each December---that is, near Halloween
and Christmas.  We examined our style clusters and found one that
contained images with red clothing and had spikes at the same two
times of the year.  This cluster was not just red, but red hats. An
assortment of winter knit beanies and red jacket hoods were present,
but what stood out were a large assortment of Santa hats as well as an
unexpected assortment of red Halloween costumes with red hats or
hoods.  It appears that red hats are uncommon enough that when they do
appear they are highly correlated with the exceptional sorts of
costumes worn on these two holidays.  Finally, the Chinese New Year
sees a similar spike in the color red in 2015 and 2016, as shown in
Figure~\ref{fig:streetstyle_chinese_new_year} (frequency of red
clothing in China).

\begin{figure}[tp]
\begin{center}
\includegraphics[width=\columnwidth]{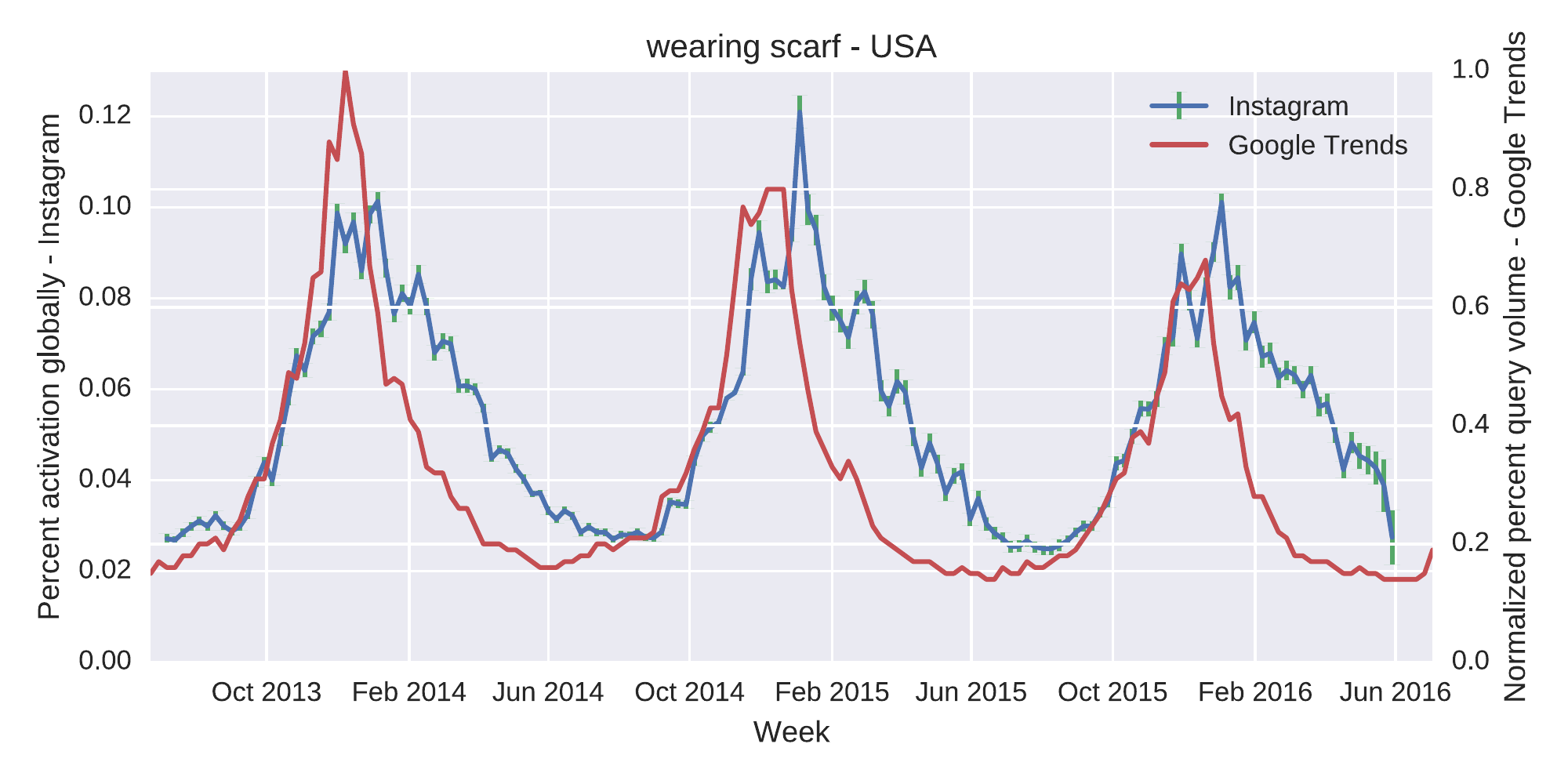}
\caption{{\bf Scarf trends in the USA.}  Our Instagram-based
  measurements are compared to Google Trends.  Instagram measurements
  are derived from mean scarf frequency aggregated by week with error
  bars indicating a 95\percent\ confidence interval.  Google Trends
  signal is per-week ``q=scarf, geo=US'' query volume normalized by
  total query volume.\label{fig:streetstyle_scarf}}
\end{center}
\end{figure}

\smallskip \noindent{\bf Clothing types.} We can also explore more
functional clothing attributes.  For example, we know people wear
scarves in cold weather, so one would expect to see a seasonal trend
in this attribute.  Figure~\ref{fig:streetstyle_scarf} shows such a
trend for visual occurrence of scarves in the United States.
Moreover, we can also compare with a secondary source of information
to estimate this signal, Google
Trends.\footnote{\url{https://www.google.com/trends/explore\#q=scarf\&geo=US}}
Google Trends provides historical trend data for Google Search
queries. By comparing these two signals, we can make several
observations.  First, the shapes of the signals are similar.  Both
feature a sudden attack and decay rather than a purely sinusoidal
shape. This correlation increases our confidence that our vision-based
signal is measuring something reasonable. Second, the onset of the
signal derived from Google Trends tends to come prior to the onset
from the Instagram photos.  One explanation for this behavior is that
people are searching for scarves on Google in order to purchase them
and Instagram allows us to measure when they are actually wearing
them. Third, while Google Trends suggests that scarves are in a
significant downward annual trend, the Instagram results serve to
constrast this trend (scarves are {\em up} in the winter of
2014-2015). While more study would be needed to determine the cause of
such a phenomenon, this comparison illustrates the utility of our
approach for devising new hypotheses for follow-up investigation. The
visual data and the Google search data combine to give us a richer
sense of what is happening in the world, and allow us to formulate new
views about the underlying human behavior.

\begin{figure}[tp]
\begin{center}
\includegraphics[width=\columnwidth]{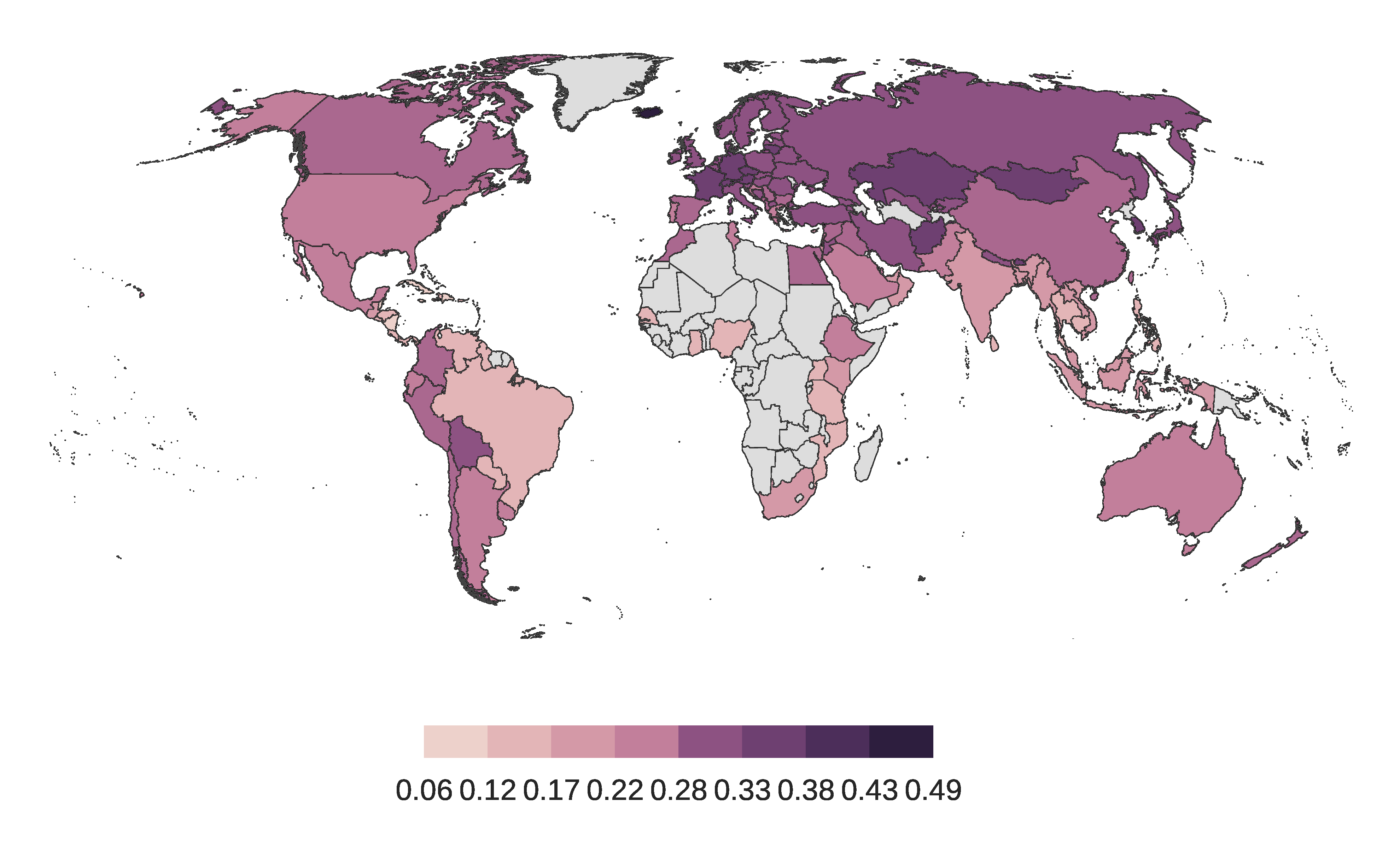}
\caption{{\bf Mean jacket frequency per country.}  Countries with $\ge$
  1,000 photos included (other countries shown in
  gray).}\label{fig:streetstyle_jackets}
\end{center}
\end{figure}

\begin{figure}[tp]
\begin{center}
\includegraphics[width=\columnwidth]{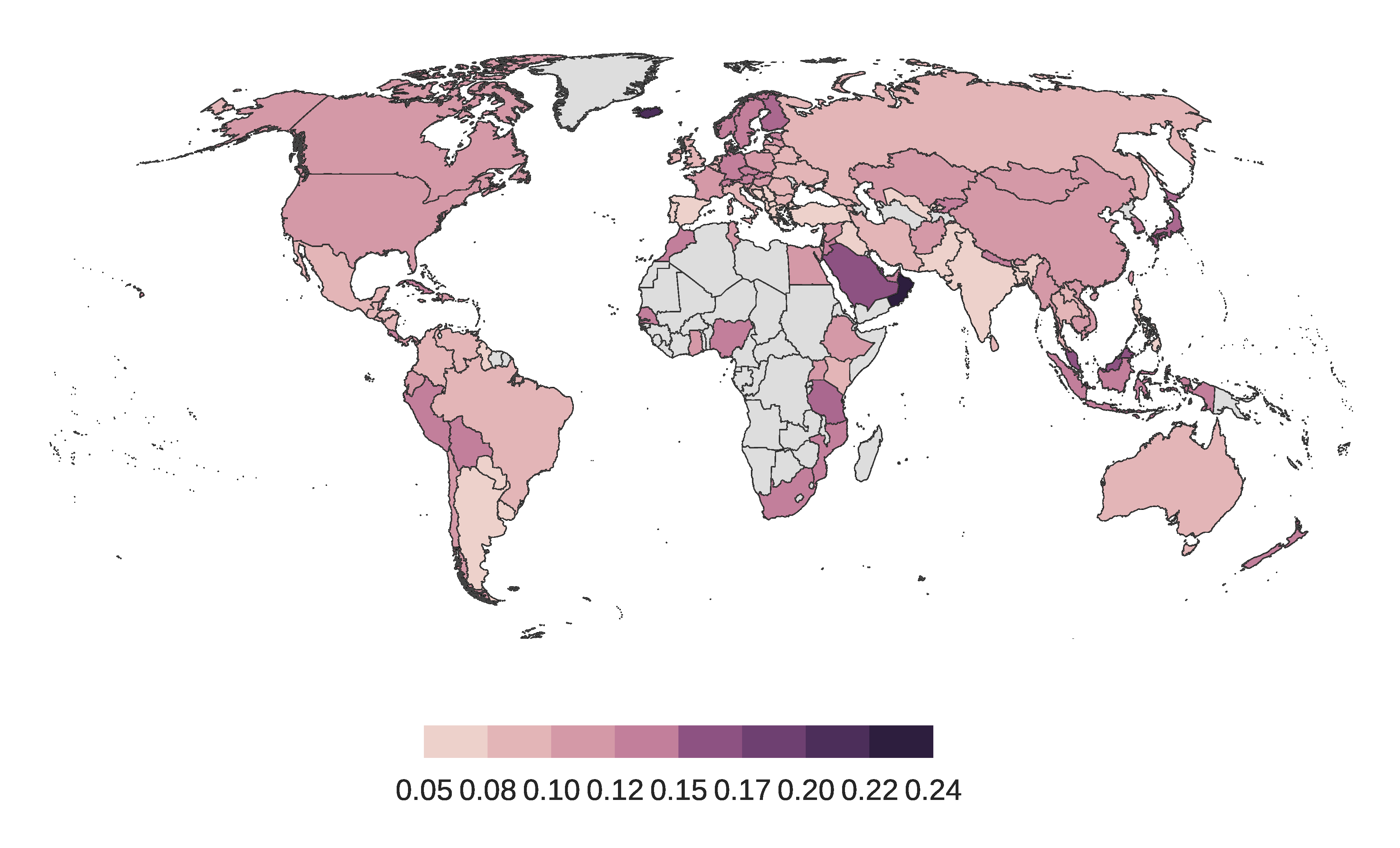}\\
\includegraphics[width=0.75\columnwidth]{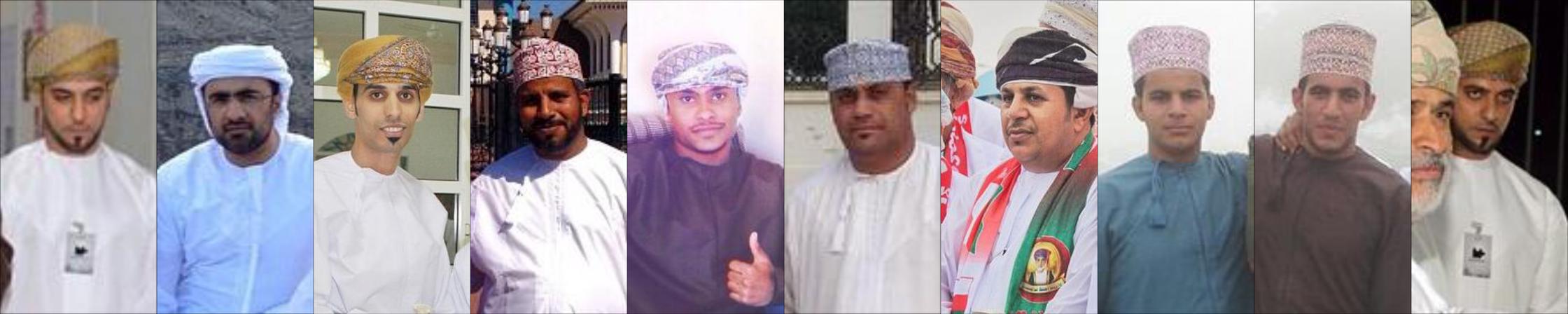}\\
\caption{Top: Mean hat frequency per country.  Countries with $\ge$ 1,000 photos
included (other countries shown in gray).  Bottom: Several top-ranked examples
of hats in Oman.  While not strictly hats, it is clear that the classifier has
learned to identify more generally, head coverings.
}\label{fig:streetstyle_hats}
\end{center}
\end{figure}

So far we have focused on temporal trends, but we can also explore
geographic trends.  Can we validate our attribute classifications
using prior knowledge of climate and weather-related clothing?  For
example, where in the world do people wear jackets?
Figure~\ref{fig:streetstyle_jackets} shows the percentage of detected
people who are wearing jackets, broken down by country.  This map
accords with intuition.  Countries further north tend to feature more
jackets.  In South America, more jackets are found as you go further
west (e.g., in Boliva or Colombia)---winter clothes will tend to be
worn more often at higher elevation (in this case in the Andes) than
at lower elevations.

What about more unexpected geographic trends?  Where in the world are
people wearing hats?  Figure~\ref{fig:streetstyle_hats} shows the
percentage of photos per country that contain people wearing hats.
Once again, people wear hats in colder places, but interestingly in
Oman, hats are evidently extremely popular.
Figure~\ref{fig:streetstyle_hats} shows several examples of images
from Oman that are highly ranked as wearing hats.  In particular, the
kuma and massar are popular in Oman, as they are an important element
of the men's national dress.

\subsection{Visualizing styles}\label{sec:vis_styles}
We also wish to visualize commonly occuring combinations of
attributes, or ``styles.'' To visualize styles, we use the method in
Section~\ref{subsec:clusters} to compute style clusters from the
embedded photos. Figure~\ref{fig:streetstyle_gmm_clusters} shows
several such clusters, in no particular order.  Given these clusters,
then for every photo in our dataset, we compute the style cluster
centroid to which it is closest.  We can then perform additional
analyses using these style clusters, such as plotting the frequency of
visual occurrence of each cluster across space and time to mine for
styles that are well localized in one or both.

\begin{figure*}[tbp]
\begin{center}
\includegraphics[width=\textwidth]{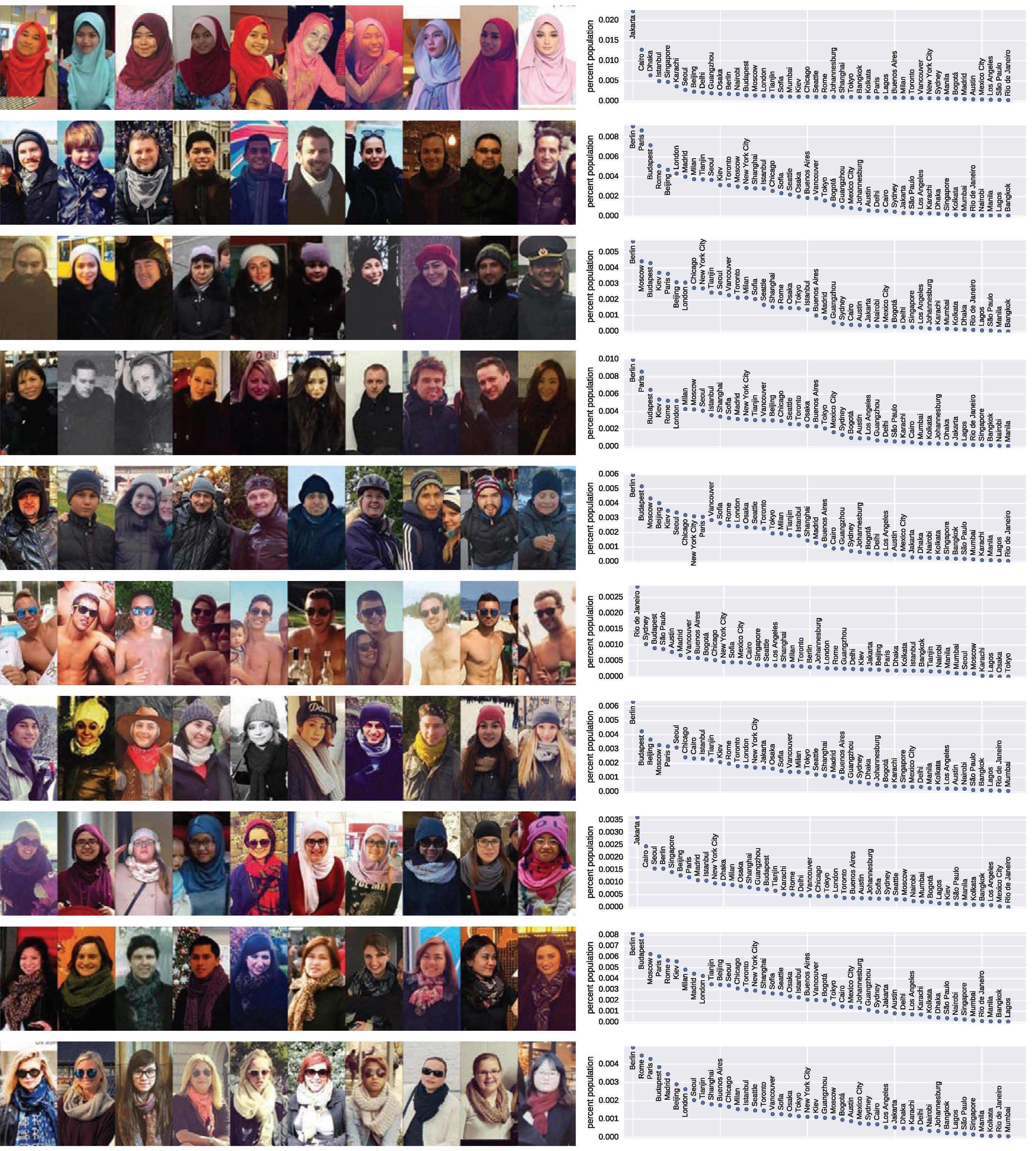}
\caption{{\bf Style clusters sorted by entropy across cities.}
  Clusters that are more distinctive for given cities are ranked
  higher. Each cluster contains thousands of people so we visually
  summarize them by the people closest to the cluster center (left),
  and the histogram of all people in the cluster across each city
  (right).\label{fig:streetstyle_rank_cities}}
\end{center}
\end{figure*}

\begin{figure*}[p]
\centering
\begin{tabular}{c}
\includegraphics[width=0.95\textwidth]{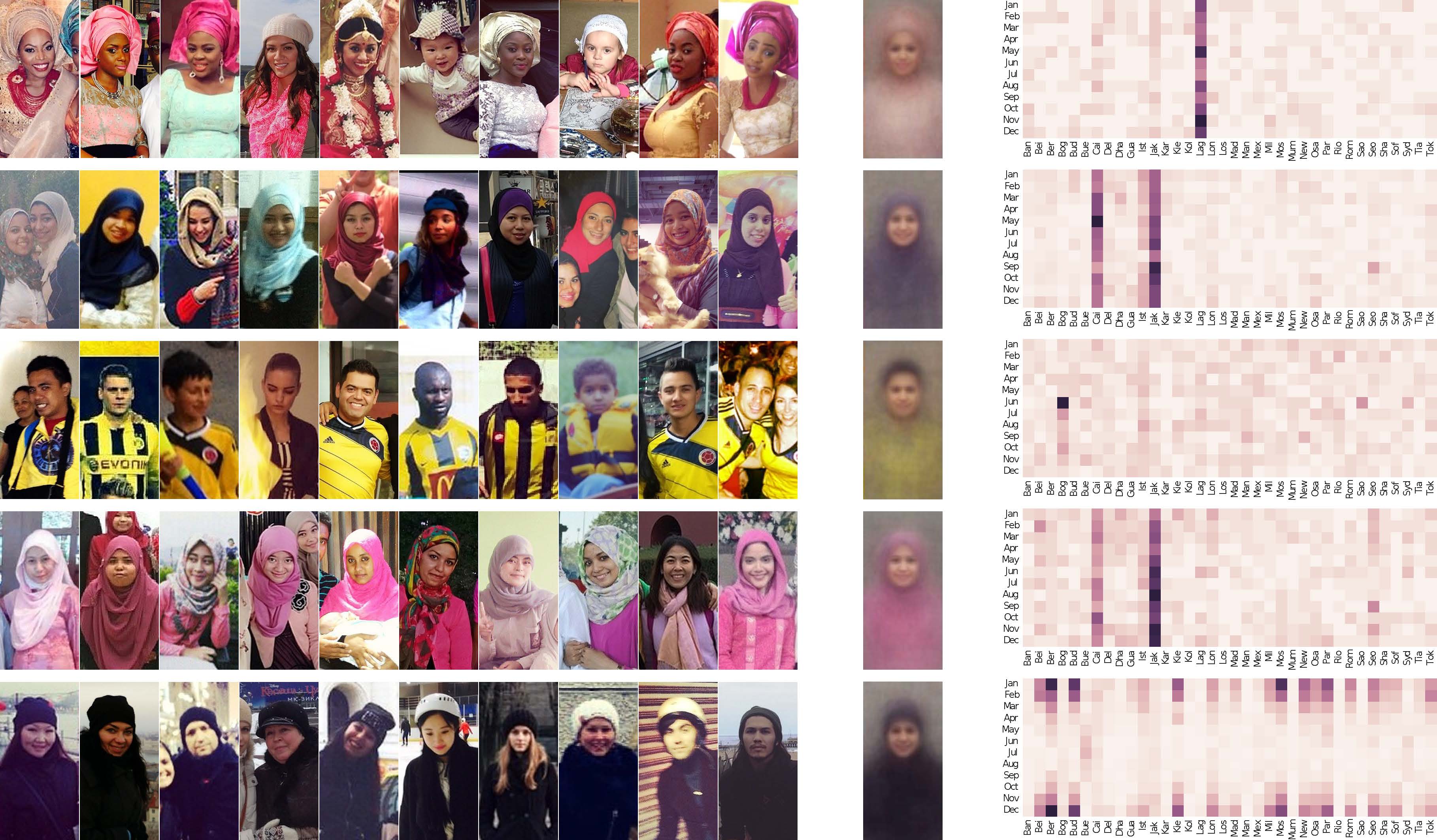}\\
Top five clusters sorted by space-time entropy.\\ \\
\includegraphics[width=0.95\textwidth]{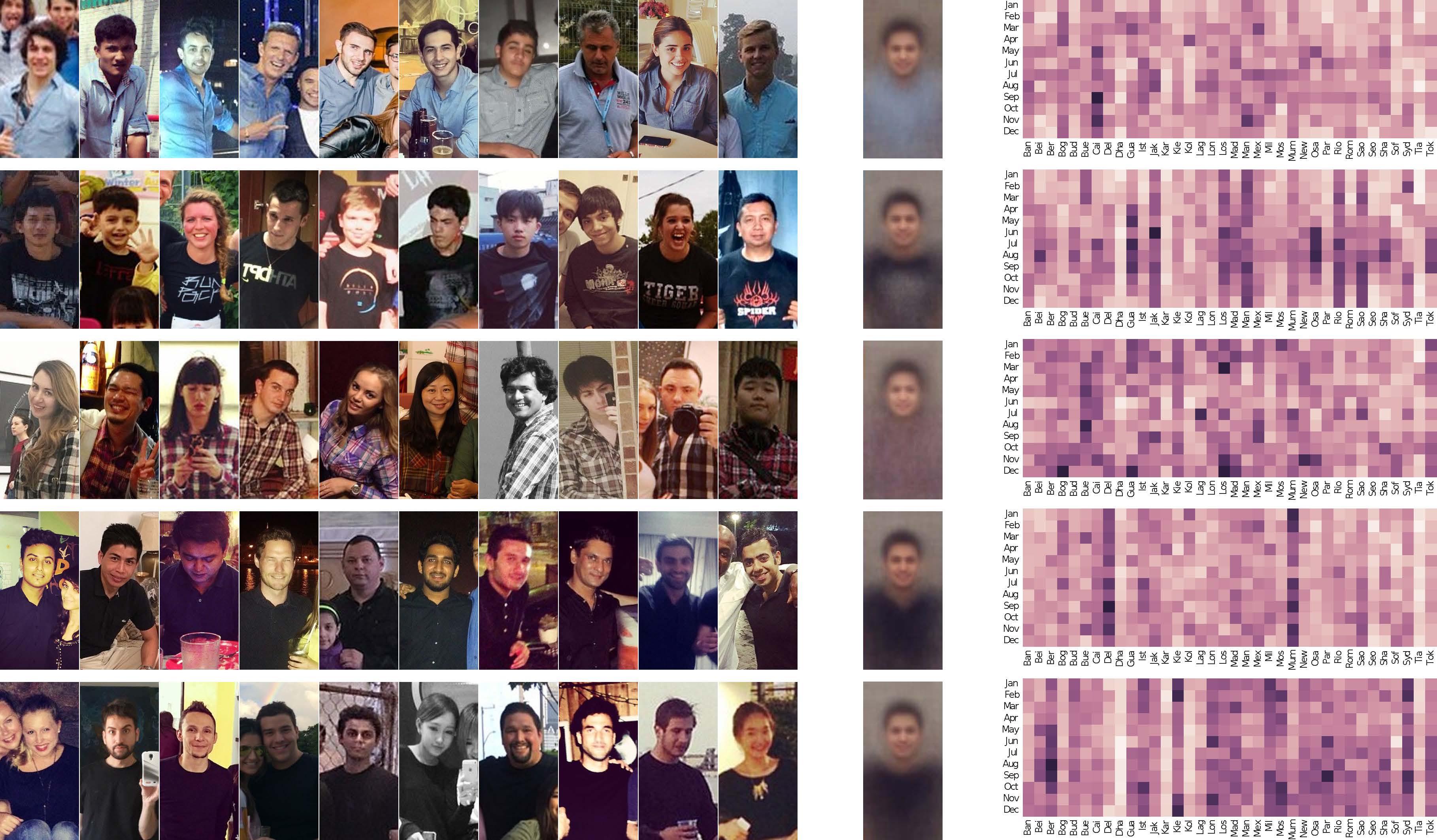}\\
Bottom five clusters sorted by space-time entropy.
\end{tabular}
\caption{{\bf Style clusters sorted by space-time entropy.} Each
  cluster contains thousands of people so we visually summarize them
  by the people closest to the cluster center (left), the average
  image of the 200 people closest to the center (middle), and the
  space-time histogram of all people in the cluster (right). The
  histogram lists 36 cities along the X-axis and months along the
  Y-axis. These histograms exhibit several types of patterns: vertical
  stripes (e.g., rows 1 and 2) indicate that the cluster appears in
  specific cities throughout the year.  Horizontal stripes correspond
  to specific months or seasons. (Note that summer and winter months
  will be reversed in the Southern Hemisphere.) For example, the fifth
  row is a winter fashion. A single bright spot can correspond to a
  special event (e.g., the soccer jerseys evident in row 3).
}\label{fig:sorted_clusters}
\end{figure*}

This raw visualization of style clusters is not particularly easy to
explore, as there are hundreds of such clusters.  Therefore, it is
useful to define a ranking function to help surface the most
interesting signals.  One way to rank clusters is according to entropy
over the different cities in the dataset---high-entropy clusters will
tend to be specific to a few cities, while low-entropy clusters tend
to be universal.  Figure~\ref{fig:streetstyle_rank_cities} shows the
top clusters for such a ranking.  This ranking helps provide a sense
of how geography affects clothing: headscarves are common in, for
instance, Jakarta and Cairo, cities in Europe tend to have a higher
percentage of winter clothing than other countries, etc.
Figure~\ref{fig:sorted_clusters} shows a set of top and bottom ranked
clusters according to entropy computed across both city and month, as
well as an average image of people in each cluster. This method for
ordering clusters is very revealing:
\begin{packed_item}
\item Regional clothing is very evident as distinct vertical
  structures in the histogram. The {\em gele} (Nigerian head-tie) is
  very distinctive of Lagos, Nigeria.
\item Yellow sports jerseys are incredibly well-concentrated at a
  specific time and place (Bogota, Colombia, during the World Cup).
\item Certain styles are common around the world and throughout the
  year. For instance, blue collared shirts, plaid shirts, and black
  t-shirts.
\end{packed_item}

\begin{figure}[h]
\begin{center}
\includegraphics[width=\columnwidth]{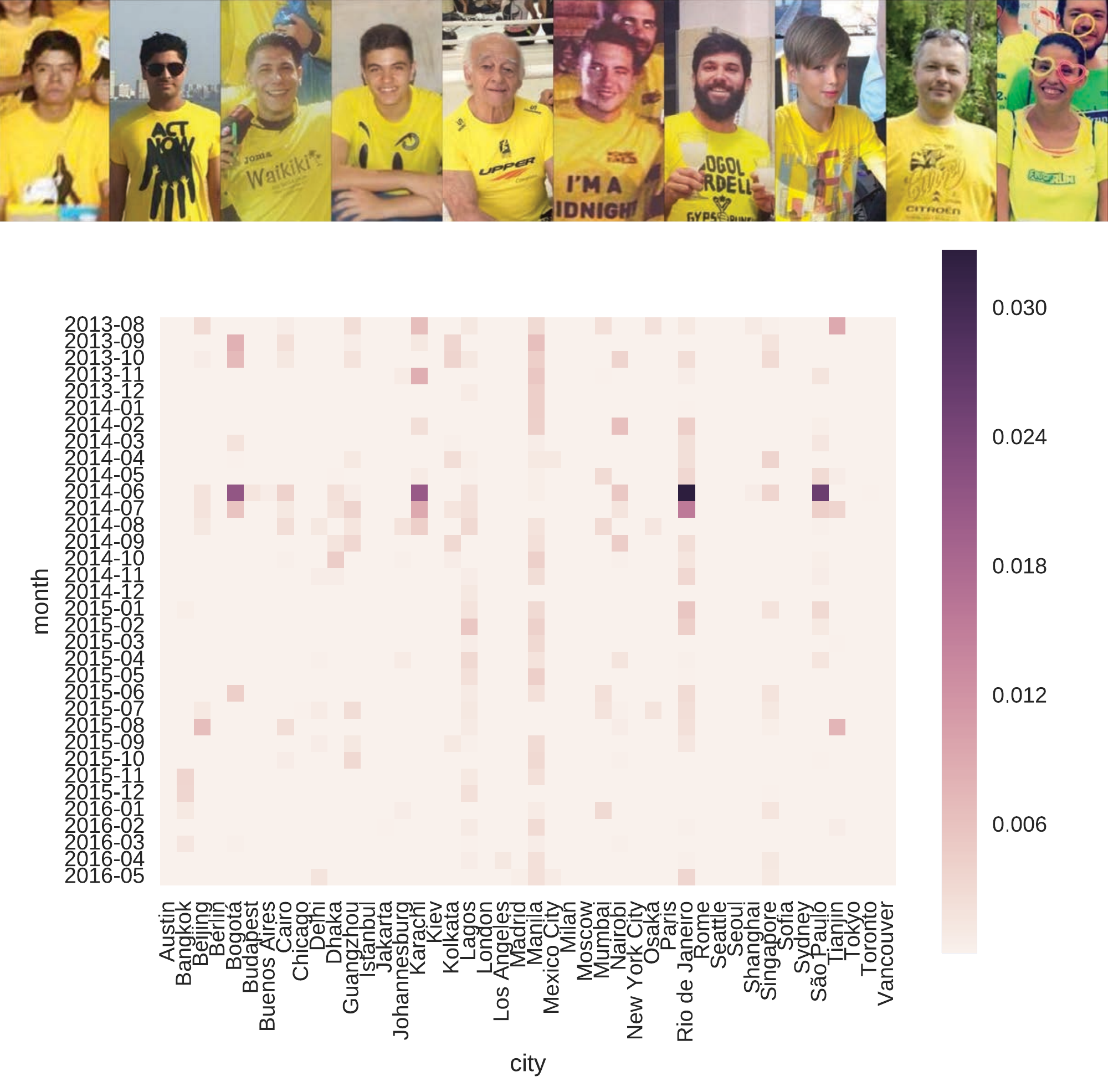}
\caption{High-ranking spatio-temporal cluster illustrating that yellow
  t-shirts with graphic patterns were very popular for a very short
  period of time, June-July 2014, in specifically Bogotá, Karachi, Rio
  de Janeiro, and São Paulo.\label{fig:streetstyle_cluster4}}
\end{center}
\end{figure}

\begin{figure}[h]
\begin{center}
\includegraphics[width=\columnwidth]{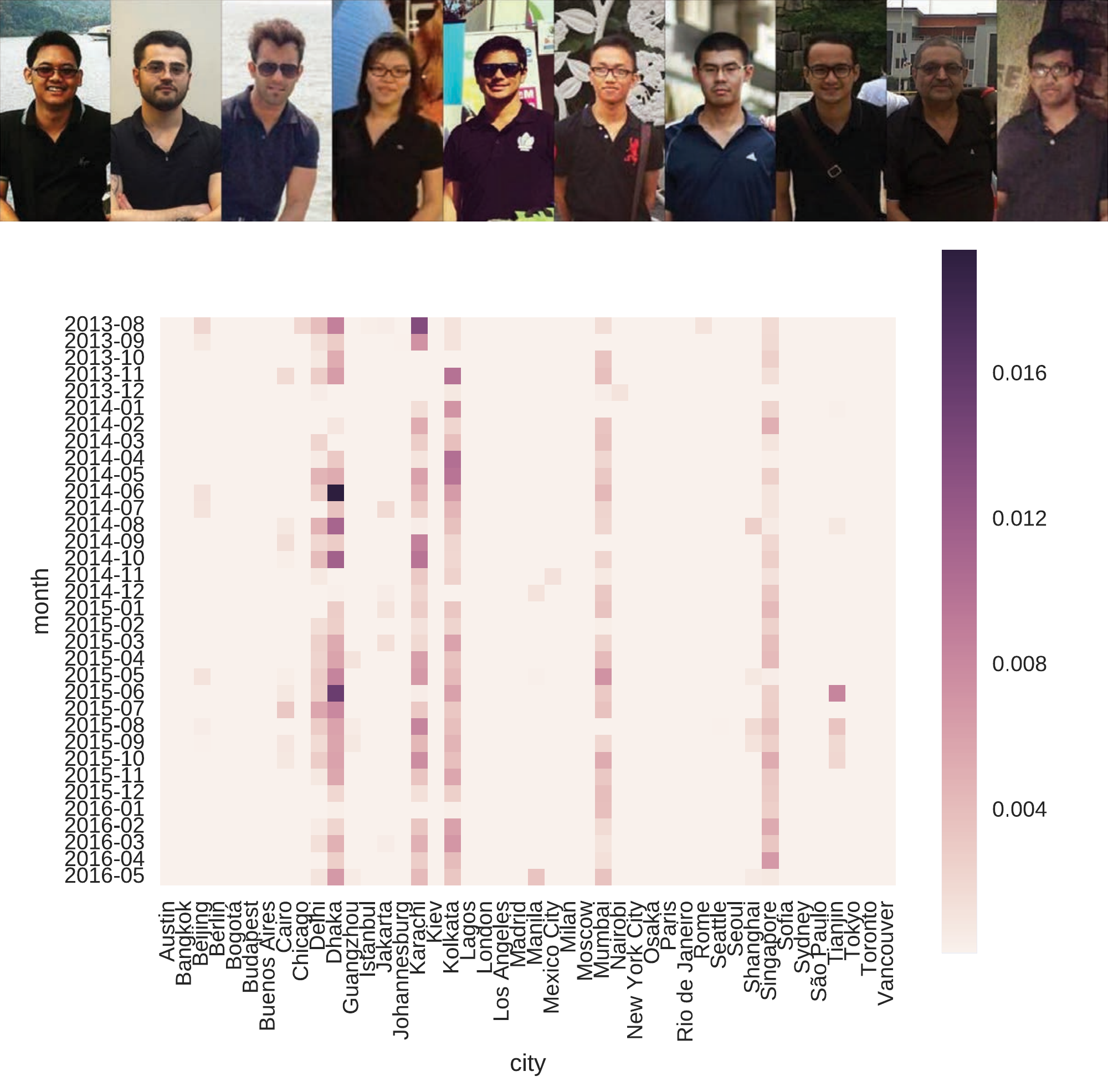}
\caption{High-ranking spatio-temporal cluster illustrating that black polo
shirts with glasses are popular in Singapore and throughout the Indian
subcontinent.\label{fig:streetstyle_cluster6}}
\end{center}
\end{figure}

Figures~\ref{fig:streetstyle_cluster4} and
\ref{fig:streetstyle_cluster6} illustrate two other interesting top
ranked clusters.  Figure~\ref{fig:streetstyle_cluster4} shows that
clusters can capture styles that become extremely popular at a very
specific time and place, and is a variant of the yellow jersey cluster
above.  In this case, June and July 2014 capture the 2014 World Cup,
and the contents of these clusters contain many examples of soccer
fans wearing team colors.  Finally,
Figure~\ref{fig:streetstyle_cluster6} shows that a style that one
might consider commonplace (black polo shirts with glasses) can be
much more popular throughout a specific region (in this case the
Indian subcontinent).

%% file: streetstyle/conclusion.tex
\section{Conclusions and Discussion}\label{sec:streetstyle_conclusion} 

In this work we presented a framework for analyzing clothing, style,
and fashion across the world using millions of photos. Towards that
end, we developed \StreetStyle, a large-scale, worldwide dataset of
people in the wild with clothing annotations, and used this dataset to
enable the analysis of a massive 15 million person photo corpus
through the application of CNNs. Our work illustrates the use of
machine learning and big data to perform visual discovery at scale.

\smallskip \noindent{\bf Limitations.} CNN-based embeddings are a very
powerful approach for organizing imagery in a high-dimensional feature
space.  Furthermore, unsupervised methods such as GMMs can be used to
further explore this space.  However, there is a limit to the
granularity of styles that we obtain. For instance, we do not learn a
clean separation between eyeglasses and sunglasses, as those were not
labeled as distinct attributes, and our style embedding does not
separate those two categories.  One way to discover finer-grained
styles would be to incorporate active learning, where users are asked
to rank similarity of styles.  Another would be to apply
weakly-supervised learning using the spatio-temporal metadata directly
in the embedding objective function rather than using it for post hoc
analysis.  Finally, our method is limited to analyzing the upper body
of the person.  As computer vision techniques for human pose
estimation mature, it would be useful to revisit this design to
normalize articulated pose for full body analysis.

\smallskip \noindent{\bf Future work.} There are many areas for future
work. We would like to more thoroughly study dataset bias. Our current
dataset measures a particular population, in a particular context
(Instagram users and the photos they take). We plan to extend to
additional image sources and compare trends across them, but also to
study aspects such as: can we distinguish between the (posed) subject
of a photo and people who may be incidentally in the background?  Does
that make a difference in the measurements?  Can we identify and take
into account differences between tourists and locals?  Are there
certain types of people that tend to be missing from such data? Do our
face detection algorithms themselves exhibit bias? We would also like
to take other difficult areas of computer vision, such as pose
estimation, and characterize them in a way that makes them amenable to
measurement at scale.

Ultimately, we would like to apply automatic data exploration
algorithms to derive, explain, and analyze the significance of
insights from machine learning (e.g., see the ``Automatic
Statistician''~\cite{lloyd:aaai:2014}). It would be also be
interesting to combine visual data with other types of information,
such as temperature, weather, and textual information on social media
to explore new types of as-yet unseen connections.  The combination of
big data, machine learning, computer vision, and automated analysis
algorithms, would make for a very powerful analysis tool more broadly
in visual discovery of fashion and many other areas.